\documentclass[10pt]{article}
\usepackage{amsmath,amsthm,amsfonts,latexsym,graphicx}
\usepackage{natbib}
\usepackage{algorithm}
\usepackage[noend]{algpseudocode}

\algrenewcommand\algorithmicthen{\relax}
\algrenewcommand\algorithmicdo{\relax}

\allowdisplaybreaks[4]

\usepackage[pdfpagemode=UseNone,pdfstartview=FitH]{hyperref}

\newtheorem{theorem}{Theorem}
\newtheorem{proposition}[theorem]{Proposition}

\theoremstyle{definition}

\theoremstyle{remark}
\newtheorem{remark}[theorem]{Remark}

\DeclareMathOperator{\Prob}{\mathbb{P}}
\DeclareMathOperator{\Expect}{\mathbb{E}}

\newcommand{\dd}{\,\mathrm{d}}
\newcommand{\ddd}{\mathrm{d}}

\newlength{\picturewidth}
\title{Retrain or not retrain: Conformal test martingales for change-point detection}
\author{Vladimir Vovk, Ivan Petej, Ilia Nouretdinov,\\Ernst Ahlberg, Lars Carlsson, and Alex Gammerman}

\begin{document}
\maketitle

\begin{abstract}
  We argue for supplementing the process of training a prediction algorithm
  by setting up a scheme for detecting the moment when the distribution of the data changes and the algorithm needs to be retrained.
  Our proposed schemes are based on exchangeability martingales, i.e., processes that are martingales under any exchangeable distribution for the data.
  Our method, based on conformal prediction, is general and can be applied on top of any modern prediction algorithm.
  Its validity is guaranteed, and in this paper we make first steps in exploring its efficiency.

   The version of this paper at \url{http://alrw.net} (Working Paper 32)
   is updated most often.
\end{abstract}

\section{Introduction}
\label{sec:introduction}

The standard assumption in mainstream machine learning is that the observed data are IID
(independent and identically distributed);
we will refer to it as the \emph{IID assumption}.
Deviations from the IID assumption are known as dataset shift,
and different kinds of dataset shift have become a popular topic of research
(see, e.g., \citet{Candela/etal:2009}).

Testing the IID assumption has been a popular topic in statistics
(see, e.g., \citet{Lehmann:2006}, Chapter 7),
but the mainstream work in statistics concentrates on the batch setting
with each observation being a real number.
In the context of deciding whether a prediction algorithm needs to be retrained,
it is more important to process data online, so that at each point in time we have an idea
of the degree to which the IID assumption has been discredited.
It is also important that the observations are not just real numbers;
in the context of machine learning the most important case is where each observation
is a pair $(x,y)$ consisting of a sample $x$ (such as an image) and its label $y$.
The existing work on detecting dataset shift in machine learning
(see, e.g., \citet{Harel/etal:2014} and its literature review)
does not have these shortcomings but does not test the IID assumption directly.

At this time the only existing method of testing the IID assumption online
is based on the method of conformal prediction \citep{Vovk/etal:2005book};
see, e.g., \citet{Vovk:2021} for a recent review.
As we explain in Section~\ref{sec:martingales},
conformal prediction allows us to construct \emph{exchangeability martingales},
which can be used as tools for testing the IID assumption.
In Section~\ref{sec:Ville} we discuss an informal scheme
that uses exchangeability martingales for deciding when a prediction algorithm needs to be retrained
and illustrate it on the well-known Wine Quality dataset.

In this paper we discuss three basic procedures for raising an alarm when the IID assumption appears to become violated.
The simplest one is the \emph{Ville procedure},
which raises an alarm when a given exchangeability martingale exceeds a given threshold.
As we explain at the beginning of Section~\ref{sec:KS},
the Ville procedure works well at the beginning of the testing process
but then becomes less efficient.
To remedy this drawback,
we introduce conformal versions of the popular CUSUM and Shiryaev--Roberts procedures
and illustrate their performance on the Wine Quality dataset.

Finally, in Section~\ref{sec:proposal} we state our proposed schemes
for deciding when a prediction algorithm should be retrained.
The two main schemes, which we call the variable and fixed schedules,
are combinations of the three basic procedures.
A short Section~\ref{sec:conclusion} concludes.

In this paper we repeatedly refer to the validity vs efficiency of our procedures.
Validity refers to their behaviour when the IID assumption is satisfied (the \emph{ideal setting});
typically, it limits the probability or frequency of false alarms.
Our procedures, being based on conformal prediction,
satisfy various properties of validity automatically.
Efficiency refers to their behaviour when the IID assumption is violated,
such as raising an alarm soon after the change point,
and achieving it is often an art.

\section{Exchangeability martingales}
\label{sec:martingales}

In this section we will define conformal test martingales,
which will be our key tool.
Let $z_1,z_2,\dots$ be an infinite sequence of observations (typically pairs $z_n=(x_n,y_n)$),
elements of a measurable space $\mathbf{Z}$, our \emph{observation space}.
An \emph{inductive conformity measure} is a measurable function $A$ mapping any observation $z\in\mathbf{Z}$
to a real number $\alpha=A(z)$, the \emph{conformity score} of $z$;
the conformity score may also depend on some prior data (in a measurable manner).
Given such an $A$, the \emph{conformal p-value} computed from observations $(z_1,\dots,z_n)\in\mathbf{Z}^*$ is
\begin{equation}\label{eq:p}
  p_n
  :=
  \frac
  {
    \left|
      \left\{
        i \mid \alpha_i<\alpha_n
      \right\}
    \right|
    +
    \theta_n
    \left|
      \left\{
        i \mid \alpha_i=\alpha_n
      \right\}
    \right|
  }
  {n},
\end{equation}
where $i$ ranges over $\{1,\dots,n\}$,
$\alpha_1=F(z_1),\dots,\alpha_n=F(z_n)$ are the conformity scores for $z_1,\dots,z_n$,
and $\theta_n$ is a random number distributed uniformly on the interval $[0,1]$.

To state the property of validity of conformal p-values in a strong form,
we need to relax the IID assumption.
For any natural number $N$, a probability measure $P$ on $\mathbf{Z}^N$ is \emph{exchangeable}
if it is invariant with respect to permutations in the following sense:
for any measurable set $E\subseteq\mathbf{Z}^N$ and any permutation $\pi$ of the set $\{1,\dots,N\}$,
\[
  P
  \left(
    (z_1,\dots,z_n) \in E
  \right)
  =
  P
  \left(
    (z_{\pi(1)},\dots,z_{\pi(n)}) \in E
  \right).
\]
The following property of validity is proved in, e.g., \citet{Vovk/etal:2005book} (Theorem~8.2).

\begin{proposition}\label{prop:validity}
  Suppose $N$ is a natural number,
  observations $z_1,\dots,z_N$ are exchangeable (i.e., generated from an exchangeable probability measure),
  $\theta_1,\dots,\theta_N$ are IID, distributed uniformly on $[0,1]$,
  and independent of the observations.
  Then the p-values $p_1,\dots,p_N$ defined by \eqref{eq:p} are IID and distributed uniformly on $[0,1]$.
\end{proposition}

A probability measure $P$ on $\mathbf{Z}^{\infty}$ over the infinite sequences $(z_1,z_2,\dots)$
is \emph{exchangeable} if, for any natural number $N$,
its restriction to the first $N$ observations $(z_1,\dots,z_N)$ is exchangeable.
According to de Finetti's representation theorem
(see, e.g., \citet{Schervish:1995}, Theorem 1.49),
assuming the observation space $\mathbf{Z}$ is a Borel space,
every exchangeable probability measure is a mixture of \emph{IID measures}
(i.e., probability measures on the infinite sequences $\mathbf{Z}^{\infty}$ of observations
under which the observations are IID).
This makes the assumptions of IID and exchangeability almost indistinguishable for $\mathbf{Z}^{\infty}$;
however, the difference is essential for finite sequences of observations.

We will test the IID assumption by testing exchangeability.
The idea is to gamble against the uniform distribution of the conformal p-values
$(p_1,p_2,\dots)\in[0,1]^{\infty}$.
A \emph{betting martingale} is a measurable function $F:[0,1]^*\to[0,\infty]$
such that $F(\Box)=1$ ($\Box$ being the empty sequence)
and, for each sequence $(u_1,\dots,u_{n-1})\in[0,1]^{n-1}$, $n\ge1$,
\begin{equation*}
  \int_0^1
  F(u_1,\dots,u_{n-1},u)
  \dd u
  =
  F(u_1,\dots,u_{n-1}).
\end{equation*}
\emph{Conformal test martingales} are defined by
\begin{equation*}
  S_n
  =
  F(p_1,\dots,p_n),
  \quad
  n=0,1,\dots,
\end{equation*}
where $p_1,p_2,\dots$ are the p-values computed by~\eqref{eq:p},
with $(\theta_1,\theta_2,\dots)$ distributed uniformly in $[0,1]^{\infty}$
and independent of the observations.

Conformal test martingales $S_n$ are bona fide martingales in the sense of satisfying
\begin{equation}\label{eq:martingale}
  \Expect(S_n \mid S_1,\dots,S_{n-1})
  =
  S_{n-1}
\end{equation}
for all $n\ge1$,
provided the observations are exchangeable,
or, as we will say, they are \emph{exchangeability martingales};
we only consider exchangeability martingales that are nonnegative and have 1 as their initial value.
We interpret $S_n$ as the amount of evidence found against our null hypothesis, the IID assumption,
after the first $n$ observations.
In betting terms, it is the capital of a tester who starts from 1 and gambles against the null hypothesis.

\begin{remark}
  Even in the case of a finite horizon, where we have $N$ observations $z_1,\dots,z_N$,
  we will have \eqref{eq:martingale} for $n=1,\dots,N$.
  The case of finite horizon is important for us since permuting a dataset
  ensures its exchangeability but not the IID assumption.
  And because of the exchangeability,
  all conformal test martingales constructed in this paper
  will always lose capital in the ideal setting.
\end{remark}

By Ville's inequality (\citet{Ville:1939}, p.~100, \citet{Shiryaev:2019}, Theorem 7.3.1),
for any constant $c>1$,
\[
  \Prob(\exists n: S_n\ge c)
  \le
  1/c.
\]
This is a property of validity for exchangeability martingales.
If, for example, we raise an alarm when $S_n$ exceeds the threshold of 100,
the probability of ever raising a false alarm will not exceed~$1\%$.

\section{Ville procedure in action}
\label{sec:Ville}

In this section we discuss a possible informal scheme for deciding when to retrain a predictor.
As an example, we consider the Wine Quality dataset \citep{Cortez/etal:2009},
available at the UCI Machine Learning repository \citep{UCI:2017}.
The dataset consists of two parts, 4898 white wines and 1599 red wines.
We randomly choose a subset of 1599 white wines and refer to it as \emph{test set 0},
and the remaining white wines (randomly permuted) will be our \emph{training set}.
All 1599 red wines form our \emph{test set 1};
therefore we have two test sets of equal sizes.

\begin{figure}
  \begin{center}
    \includegraphics[width=\picturewidth]{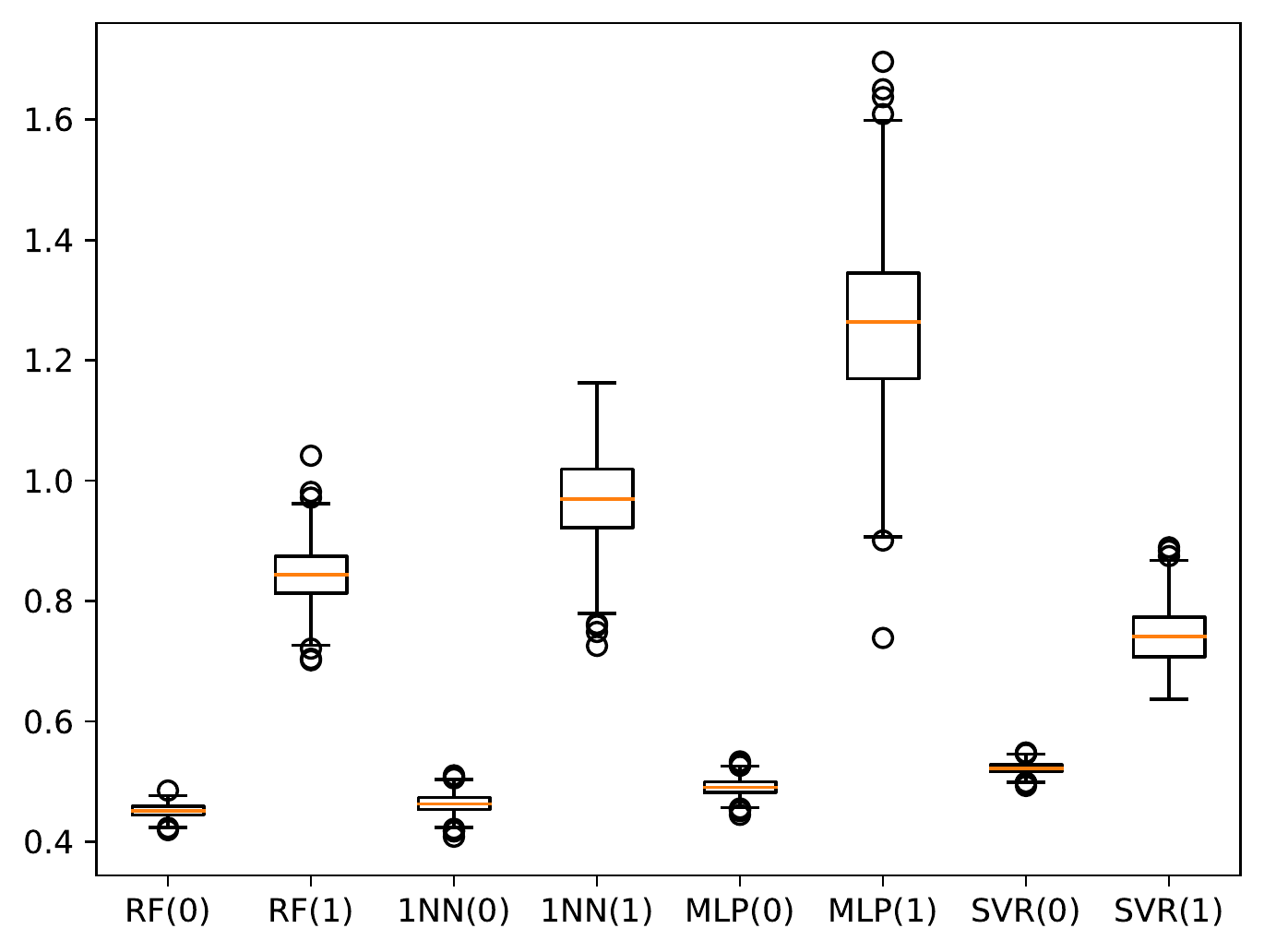}
  \end{center}
  \caption{The accuracies of various prediction algorithms on the Wine Quality dataset,
    as described in text}
  \label{fig:wine-accuracies}
\end{figure}

We will be interested in two scenarios.
In scenario 0 we train our prediction algorithm on the full training set
and test the resulting model on test set 0.
We can expect the quality of prediction to be good,
since the training and test set are coming from the same distribution.
If we normalize the data by using \texttt{StandardScaler} in \texttt{scikit-learn} \citep{scikit-learn:2011},
we can achieve the test MAD (mean absolute deviation) of about 0.45.
The best values achieved by the algorithms implemented in \texttt{scikit-learn}
for the default values of the parameters are given in Figure~\ref{fig:wine-accuracies},
where RF stands for Random Forest, 1-NN for 1-Nearest Neighbour, 
MLP for Multilayer Perceptron, and SVR for Support Vector Regression.
The relevant boxplots are those marked with 0 in parentheses;
the boxplots are over 1000 simulations and shown to give an idea of the dependence
on the seed used for the random number generator
(the seed affects the split into the training and test sets
and may be used internally by the prediction algorithm, e.g., by Random Forest).
The algorithms are ordered by their performance in scenario 0.
In scenario 1 we test the same trained model on test set 1;
since its distribution is different (from the very start of the test set),
the resulting test MAD will be significantly worse,
as indicated in Figure~\ref{fig:wine-accuracies} by the boxplots marked with 1 in parentheses.

To detect a possible change point in the test set (which does not exist in test set 0 and is the very start in test set 1),
the training set of 3299 white wines is randomly split into three \emph{folds} of nearly equal sizes,
1100, 1100, and 1099.
We use each fold in turn as the \emph{calibration set} and the remaining folds as the \emph{training set proper}.
For each fold $k\in\{1,2,3\}$ we train a prediction algorithm on the training set proper
and run an exchangeability martingale (based, in some way, on the resulting model)
on the $1100+1599=2699$ observations $z'_1,\dots,z'_{1100},z''_1,\dots,z''_{1599}$,
where $z'_1,\dots,z'_{1100}$ is the calibration set and $z''_1,\dots,z''_{1599}$ is the test set
(for one of the folds, 1100 should be replaced by 1099, but we will ignore this in our discussion).
This way we obtain three paths, plots of the values of the exchangeability martingales vs time.
We still have two scenarios:
scenario 0 uses test set 0, and scenario 1 uses test set 1;
thus we have 6 paths overall.

\begin{figure}
  \begin{center}
    \includegraphics[width=\picturewidth]{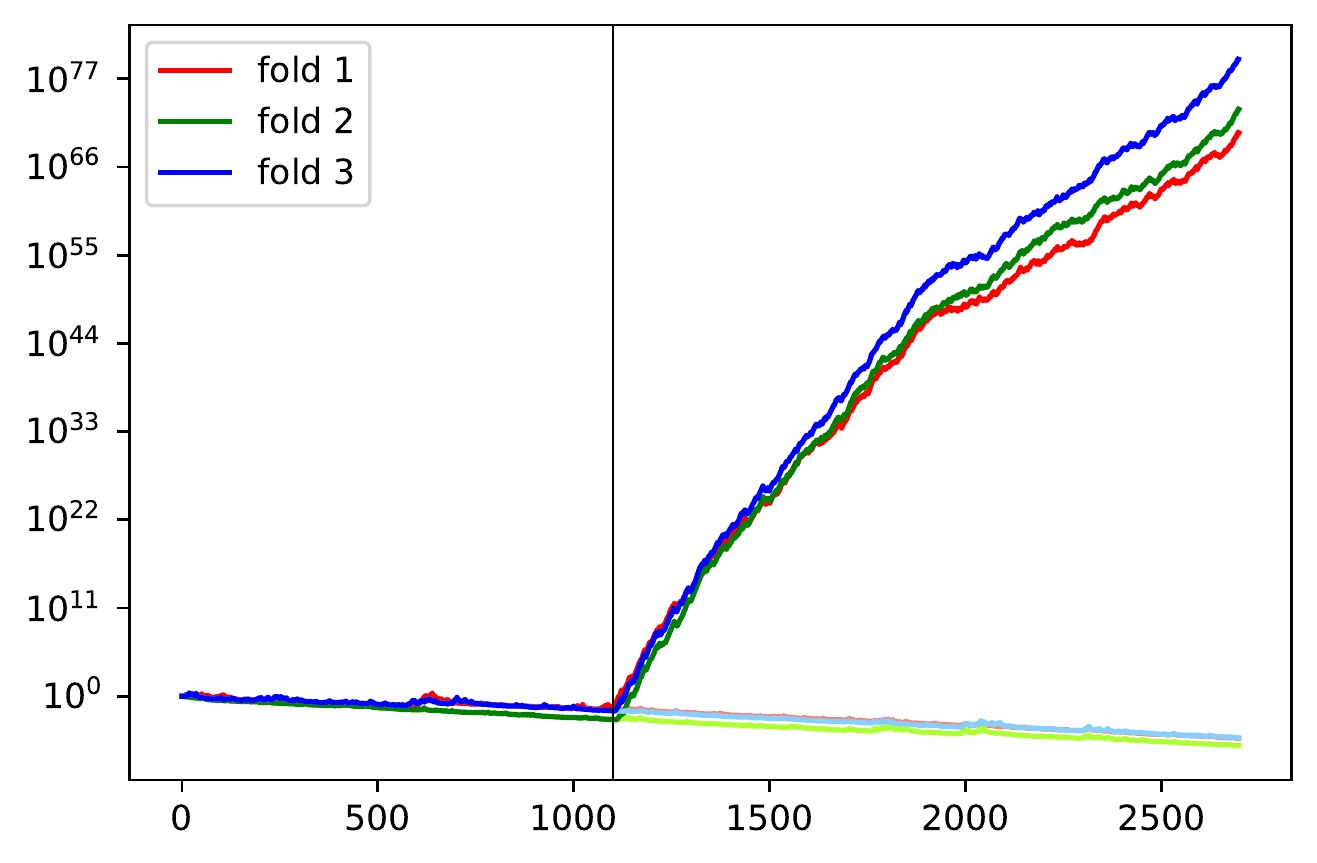}
  \end{center}
  \caption{Paths of the three conformal test martingales
    (based on Random Forest, conformity measure $y-\hat y$, and Simple Jumper)
    applied to the Wine Quality dataset as described in text}
  \label{fig:wine-RFdiff-Ville-full}
\end{figure}

Results for a specific conformity measure (based on Random Forest) and betting martingale (to be defined shortly)
are shown in Figure~\ref{fig:wine-RFdiff-Ville-full}.
For each fold we have a conformal test martingale,
and paths of these three martingales are shown using different colours,
as indicated in the legend.
We have two paths for each of the martingales:
the one over the calibration set and test set 0 (with the part over test set 0 shown in lighter colours),
and the other over the calibration set and test set 1.
The behaviour of the three martingales in scenario 1 is similar,
all achieve a high value, of the order of magnitude about $10^{70}$,
and start rising sharply soon after the change point (shown as the thin vertical line);
the presence of the change point becomes obvious shortly after it happens.

The conformity measure used in Figure~\ref{fig:wine-RFdiff-Ville-full} is
\begin{equation}\label{eq:diff}
  \alpha_i
  :=
  y_i - \hat y_i,
\end{equation}
where $\hat y_i$ is the prediction for the label $y_i$ of the sample $x_i$
produced by the model found from the training set proper.

\begin{algorithm}[bt]
  \caption{Simple Jumper ($(p_1,p_2,\dots)\mapsto(S_1,S_2,\dots)$)}
  \label{alg:SJ}
  \begin{algorithmic}[1]
    \State $C_{-1}:=C_0:=C_1:=1/3$
    \State $C:=1$
    \For{$n=1,2,\dots$:}
      \For{$\epsilon\in\{-1,0,1\}$:}
        $C_{\epsilon} = (1-J)C_{\epsilon} + (J/3)C$
      \EndFor
      \For{$\epsilon\in\{-1,0,1\}$:}
        $C_{\epsilon} = C_{\epsilon} f_{\epsilon}(p_n)$
      \EndFor
      \State $S_n := C := C_{-1}+C_0+C_1$
    \EndFor
  \end{algorithmic}
\end{algorithm}

To transform the p-values $p_1,p_2,\dots$ computed by \eqref{eq:p}
into a conformal test martingale
we use the betting martingale
\[
  F(p_1,\dots,p_n)
  :=
  \int
  \left(
    \prod_{i=1}^n
    f_{\epsilon_i}(p_i)
  \right)
  \mu(\ddd(\epsilon_0,\epsilon_1,\dots)),
\]
where
\begin{equation}\label{eq:f}
  f_{\epsilon}(p)
  :=
  1 + \epsilon(p-0.5)
\end{equation}
and $\mu$ is the following Markov chain with state space $\{-1,0,1\}$:
the initial state is $\epsilon_0=-1,0,1$ with equal probabilities,
and the transition function prescribes maintaining the same state with probability $1-J$
and, with probability $J$, choosing a random state from the state space $\{-1,0,1\}$.
The intuition is that at each step $i$  we are using one of the \emph{betting functions} \eqref{eq:f};
$f_{-1}$ corresponds to betting on small values of $p_i$,
$f_{1}$ corresponds to betting on large values of $p_i$,
and $f_0$ corresponds to not betting.
We always set $J:=0.01$.
Therefore, we start from the uniform allocation of the initial capital to the states
and usually continue betting in the same way as on the previous step.
We will refer to this betting martingale as the \emph{Simple Jumper}
(it is a simplification of the Sleepy Jumper described in \citet{Vovk/etal:2005book}, Section~7.1).
The pseudocode for the Simple Jumper applied to the p-values \eqref{eq:p} is given as Algorithm~\ref{alg:SJ}.

\begin{figure}
  \begin{center}
    \includegraphics[width=\picturewidth]{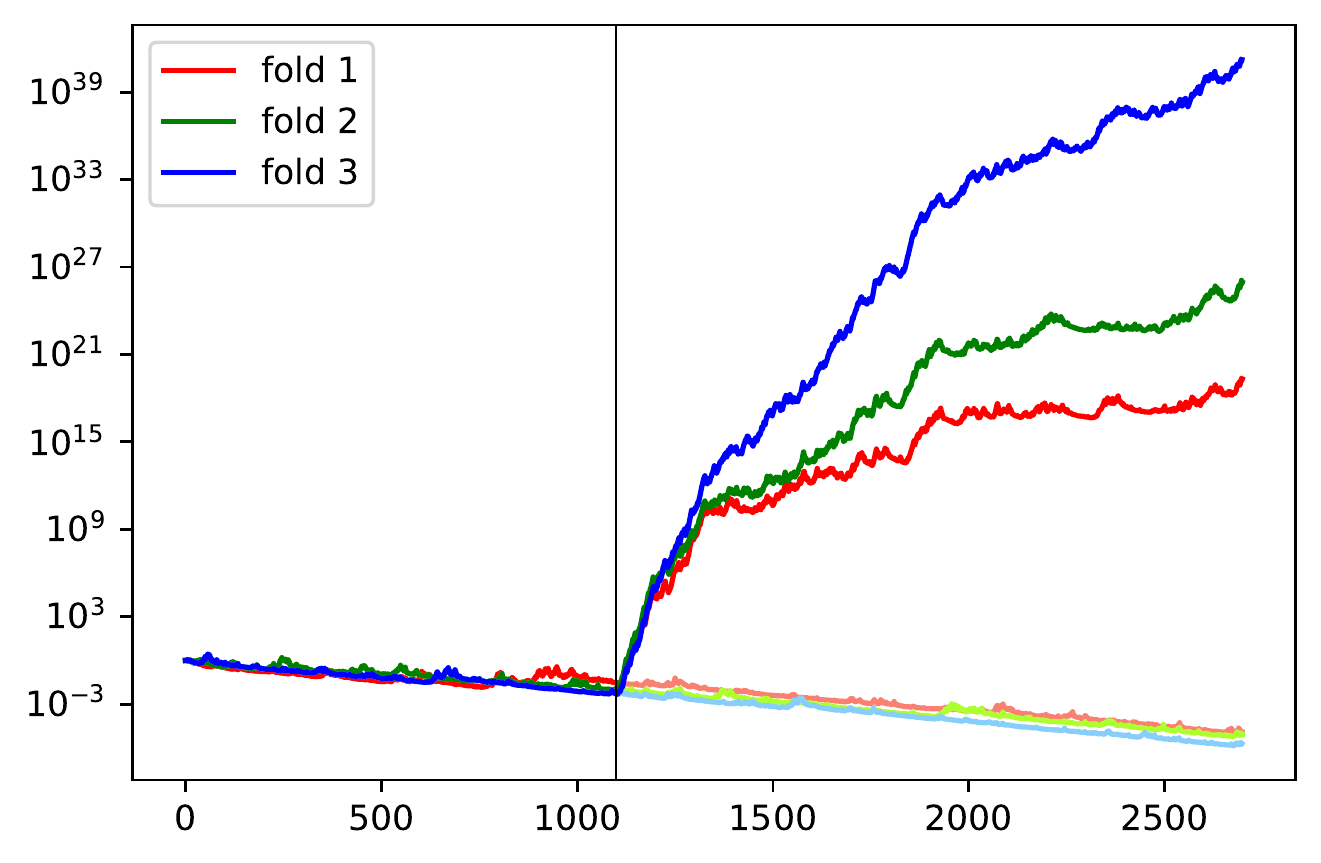}
  \end{center}
  \caption{Analogue of Figure~\ref{fig:wine-RFdiff-Ville-full}
    with the conformity measure $y - \hat y$ replaced by $\left|y - \hat y\right|$}
  \label{fig:wine-RFabsdiff-Ville-full}
\end{figure}

Replacing the conformity measure \eqref{eq:diff} by its absolute value,
\begin{equation}\label{eq:absdiff}
  \alpha_i
  :=
  \left|y_i - \hat y_i\right|,
\end{equation}
leads to a slower growth, as illustrated in Figure~\ref{fig:wine-RFabsdiff-Ville-full},
and in this paper we concentrate on the signed version \eqref{eq:diff}.
In the case of fold~1 (the red line),
we can see a pronounced phenomenon of ``decay'' setting in around observation 2000;
as it were, the new distribution (corresponding to red wines) becomes a new normal,
and the growth of the martingale stops.

\begin{figure}
  \begin{center}
    \includegraphics[width=\picturewidth]{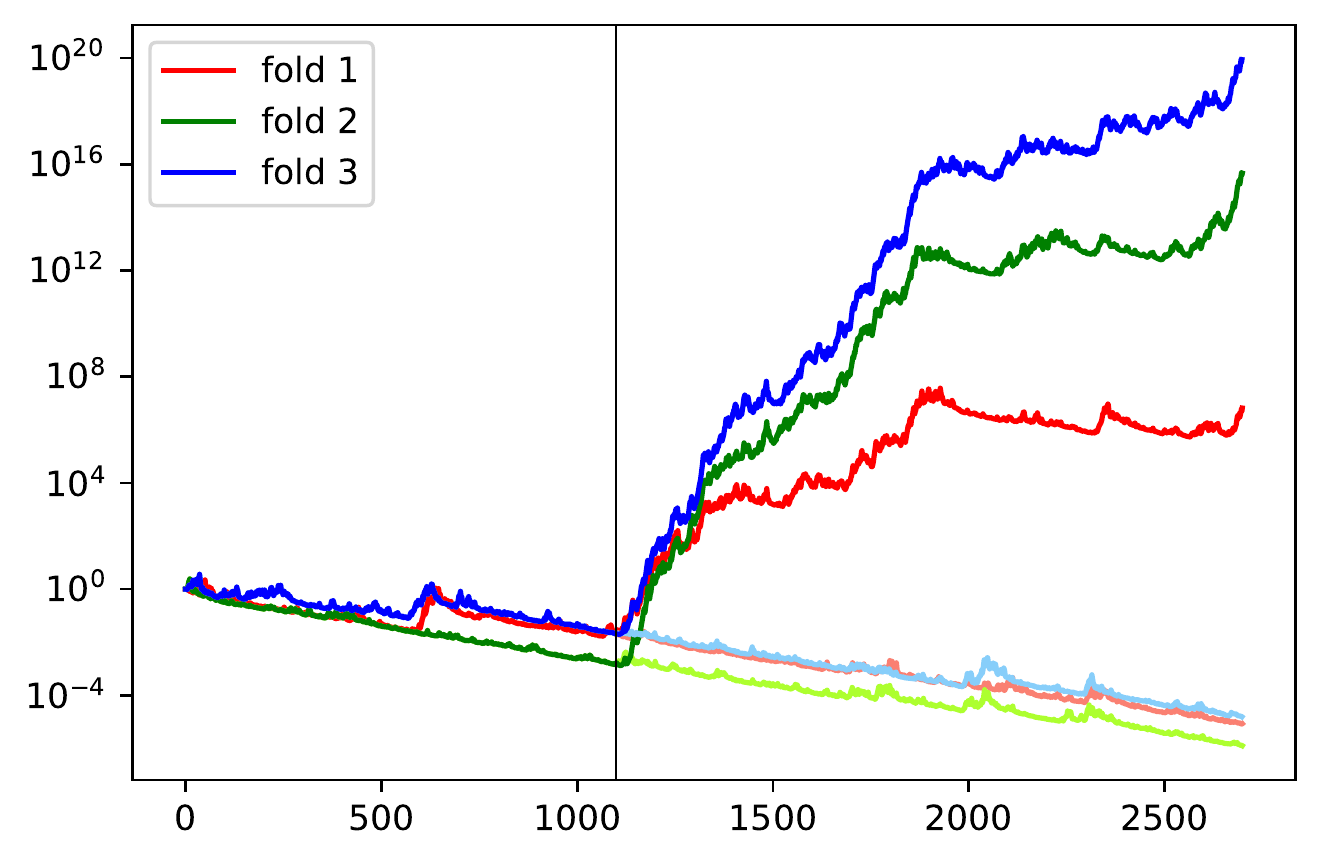}
  \end{center}
  \caption{Analogue of Figure~\ref{fig:wine-RFdiff-Ville-full}
    for the PIT conformity measure $F_i(y_i)$ (with $F_i$ produced by Random Forest)}
  \label{fig:wine-RFpit-Ville-full}
\end{figure}

In the case of Random Forest, there is an interesting alternative to the conformity measure \eqref{eq:diff}.
With the default values of the parameters in \texttt{scikit-learn},
its prediction $\hat y_i$ is computed by averaging the predictions
produced by 100 decision trees.
The \emph{PIT conformity measure} (where PIT stands for ``probability integral transform'')
is $\alpha_i:=F_i(y_i)$, where $F_i$ is the empirical distribution function
determined by the predictions $\hat y_i^j$ produced by the decision trees $j=1,\dots,100$.
In other words,
\[
  \alpha_i
  :=
  \left|\left\{
    j\mid \hat y_i^j \le y_i
  \right\}\right|
  / 100.
\]
The results are shown in Figure~\ref{fig:wine-RFpit-Ville-full};
the growth of the conformal test martingales in scenario 1
becomes even weaker than for the conformity measure~\eqref{eq:absdiff}.

\begin{figure}
  \begin{center}
    \includegraphics[width=\picturewidth]{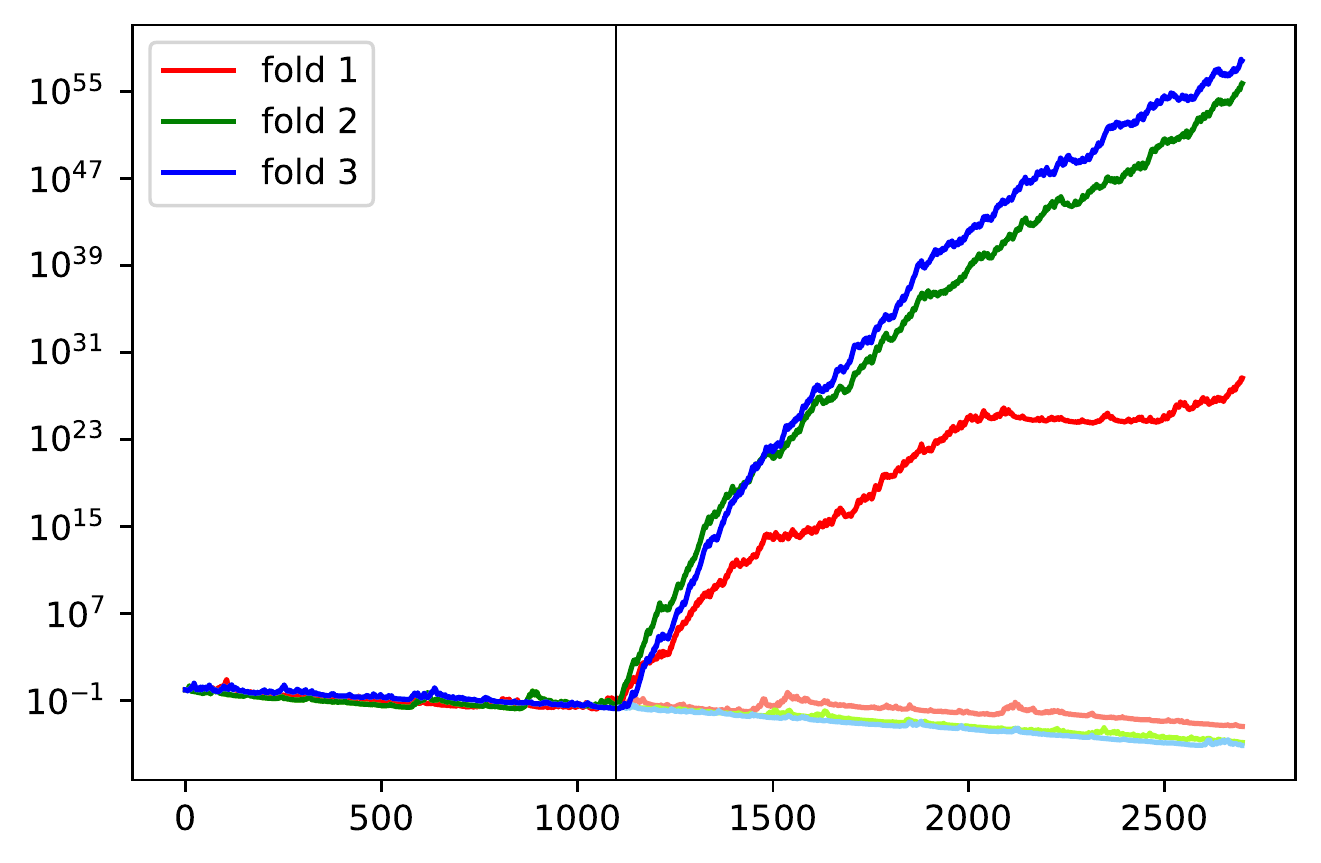}
  \end{center}
  \caption{Analogue of Figure~\ref{fig:wine-RFdiff-Ville-full}
    with $\hat y$ produced by 1-Nearest Neighbour}
  \label{fig:wine-NNdiff-Ville-full}
\end{figure}

\begin{figure}
  \begin{center}
    \includegraphics[width=\picturewidth]{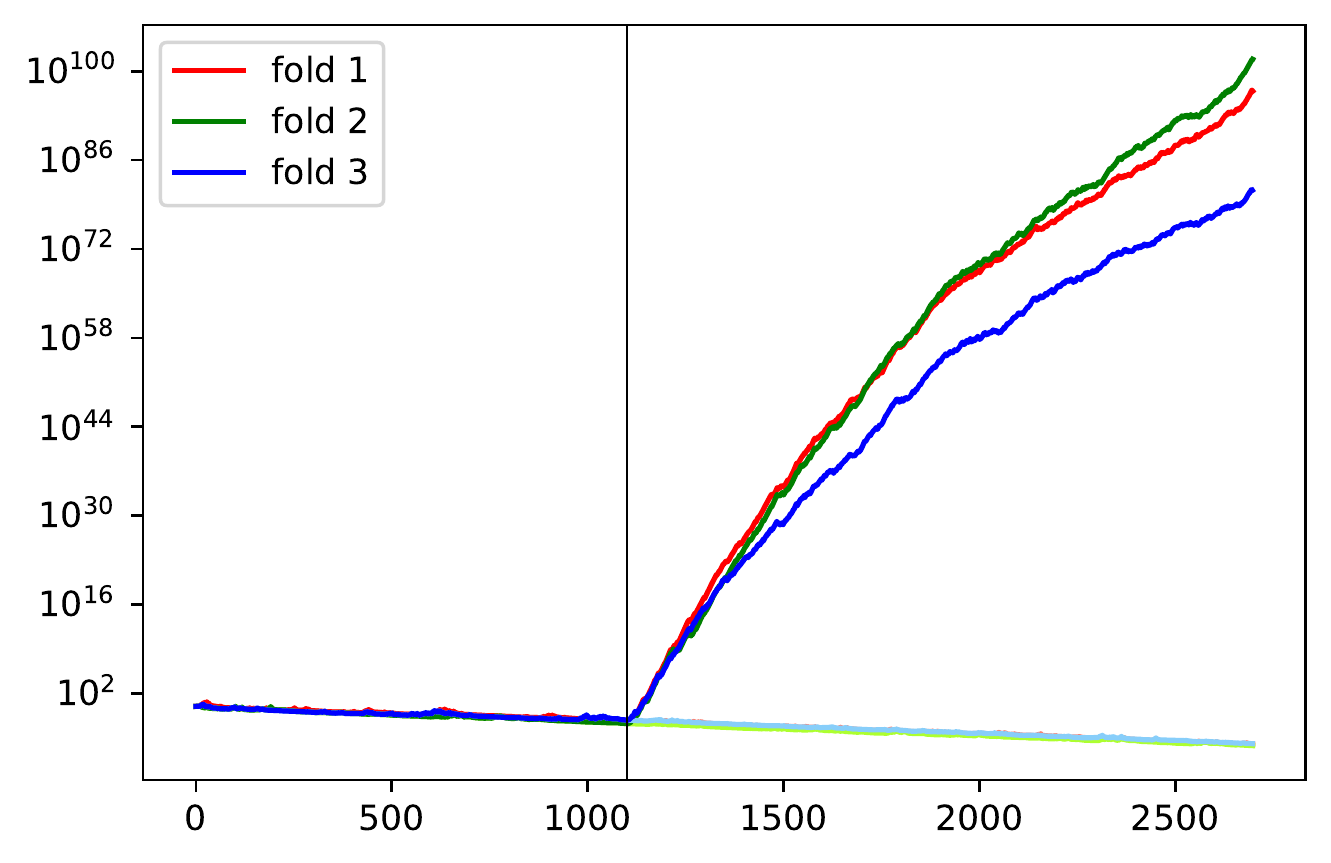}
  \end{center}
  \caption{Analogue of Figure~\ref{fig:wine-RFdiff-Ville-full}
    with $\hat y$ produced by Multilayer Perceptron}
  \label{fig:wine-MLPdiff-Ville-full}
\end{figure}

\begin{figure}
  \begin{center}
    \includegraphics[width=\picturewidth]{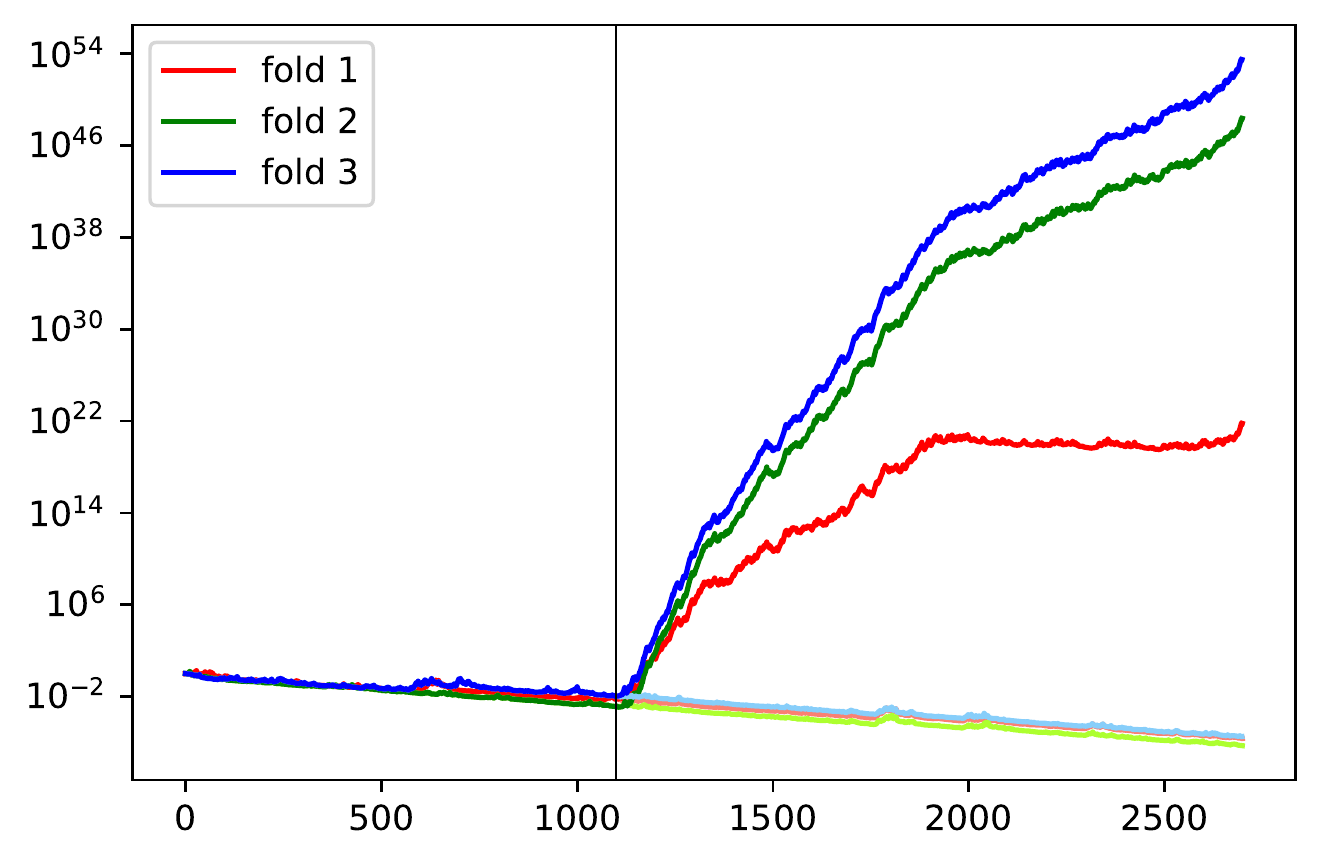}
  \end{center}
  \caption{Analogue of Figure~\ref{fig:wine-RFdiff-Ville-full}
    with $\hat y$ produced by Support Vector Regression}
  \label{fig:wine-SVRdiff-Ville-full}
\end{figure}

The conformity measure \eqref{eq:diff} can be applied to any regressor.
Figures~\ref{fig:wine-NNdiff-Ville-full}, \ref{fig:wine-MLPdiff-Ville-full}, and~\ref{fig:wine-SVRdiff-Ville-full}
are the analogues of Figure~\ref{fig:wine-RFdiff-Ville-full}
for 1-Nearest Neighbour, Multilayer Perceptron, and Support Vector Regression.
Notice that our martingales detect lack of exchangeability best
in situations where it matters most:
according to Figure~\ref{fig:wine-accuracies}, the accuracy of Multilayer Perceptron
suffers most of the dataset shift,
and according to Figure~\ref{fig:wine-MLPdiff-Ville-full},
the conformal test martingales based on this algorithm achieve the fastest growth.
In some cases (as for fold~1 in Figure~\ref{fig:wine-SVRdiff-Ville-full})
the phenomenon of decay is even more pronounced than in Figure~\ref{fig:wine-RFabsdiff-Ville-full}.

The procedure described in this section may be used for deciding when to retrain:
e.g., we may decide to retrain when one of the three martingales
exceeds the threshold 100.
In this case,
the probability of ever raising a false alarm never exceeds~3\%.

\subsection*{Detecting covariate shift}

The conformity measures that we have used so far in this section
were functions of the true labels and predictions.
If the underlying algorithm is robust to moderate covariate shift,
the resulting testing procedures will not detect deviations from exchangeability
under such covariate shift.
It makes sense since no retraining is required in this case.

\begin{figure}
  \begin{center}
    \includegraphics[width=\picturewidth]{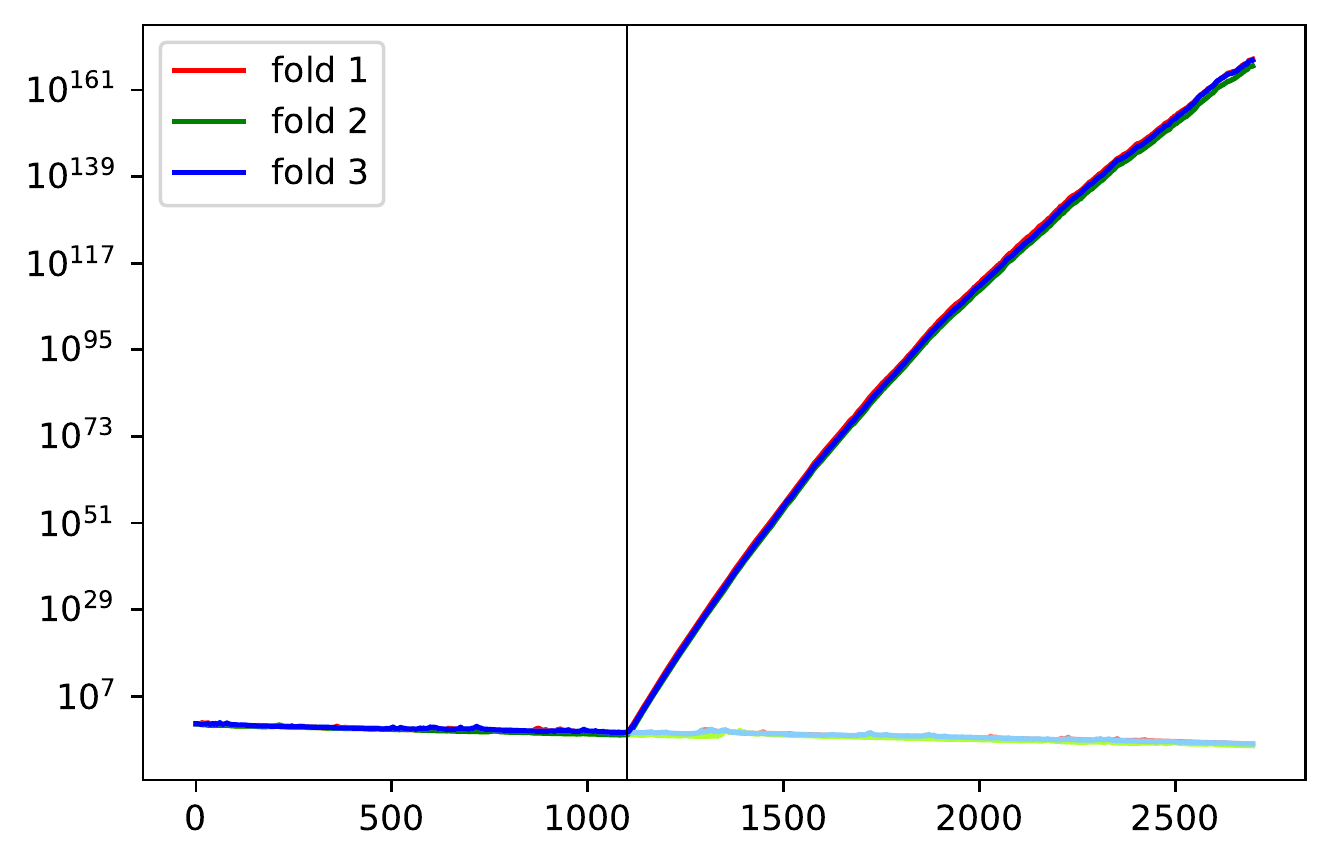}
  \end{center}
  \caption{Paths of the three conformal test martingales
    based on the nearest-distance conformity measure
    applied to the Wine Quality dataset}
  \label{fig:wine-IND-Ville-full}
\end{figure}

Figure~\ref{fig:wine-IND-Ville-full} shows the results
for the conformity score $\alpha$ of an observation $(x,y)$
(in the calibration or test set)
computed as the distance from $x$ to the nearest sample in the training set proper.
This conformity measure completely ignores the labels
but still achieves spectacular values for conformal test martingales based on it.

\section{CUSUM and Shiryaev--Roberts procedures for change detection}
\label{sec:KS}

A well-known disadvantage of the Ville procedure
is that it becomes less and less efficient as time passes and the value of the martingale goes down,
which inevitably happens in the absence of change points:
cf.\ scenario 0 (lighter colours) in Figures~\ref{fig:wine-RFdiff-Ville-full}--\ref{fig:wine-IND-Ville-full}.
It may take a long time to recover the lost capital and so to detect the change point.
We can say that, whereas the Ville procedure may be suitable during the first stages of the testing process
(while our capital is still not negligible),
it is less suitable for later stages.

Let $S$ be an exchangeability martingale that never takes value 0
(such are all martingales considered in the previous section).
The \emph{CUSUM procedure} \citep{Page:1954} raises an alarm at the time
\begin{equation}\label{eq:CUSUM}
  \tau
  :=
  \min
  \left\{
    n \mid \gamma_n \ge c
  \right\},
  \text{ where }
  \gamma_n
  :=
  \max_{i=0,\dots,n-1} \frac{S_n}{S_i}
\end{equation}
and $c>1$ is the parameter of the procedure.
The Shiryaev--Roberts procedure \citep{Shiryaev:1963,Roberts:1966}
modifies this by replacing the maximum with a sum:
\begin{equation}\label{eq:SR}
  \sigma
  :=
  \min
  \left\{
    n \mid \psi_n \ge c
  \right\},
  \text{ where }
  \psi_n
  :=
  \sum_{i=0}^{n-1} \frac{S_n}{S_i}.
\end{equation}
In both \eqref{eq:CUSUM} and \eqref{eq:SR}, $\min\emptyset:=\infty$.
The maxima $\gamma_n$ in \eqref{eq:CUSUM} (with $\gamma_0:=0$) are called the \emph{CUSUM statistics},
and the sums $\psi_n$ (with $\psi_0:=0$) in \eqref{eq:SR} are called the \emph{Shiryaev--Roberts statistics}.

It is known that, under the null hypothesis (the IID assumption in this context),
$\Expect(\sigma)\ge c$
(see, e.g., \citet{Vovk:2021}, Proposition 4.2);
moreover, $\Expect(\sigma)$ equals $c$
if we ignore the possibility of the Shiryaev--Roberts statistics overshooting the threshold $c$.
Since $\tau\ge\sigma$, we also have $\Expect(\tau)\ge c$.
It is also shown in \citet{Vovk:2021} (Proposition 4.4) that,
when the Shiryaev--Roberts procedure is applied repeatedly,
the relative frequency of false alarms will not exceed $1/c$ in the long run
(this statement is informative only when $\Expect(\sigma)=\infty$,
since otherwise it follows from $\Expect(\sigma)\ge c$ and Kolmogorov's strong law of large numbers).

\begin{figure}
  \begin{center}
    \includegraphics[width=\picturewidth]{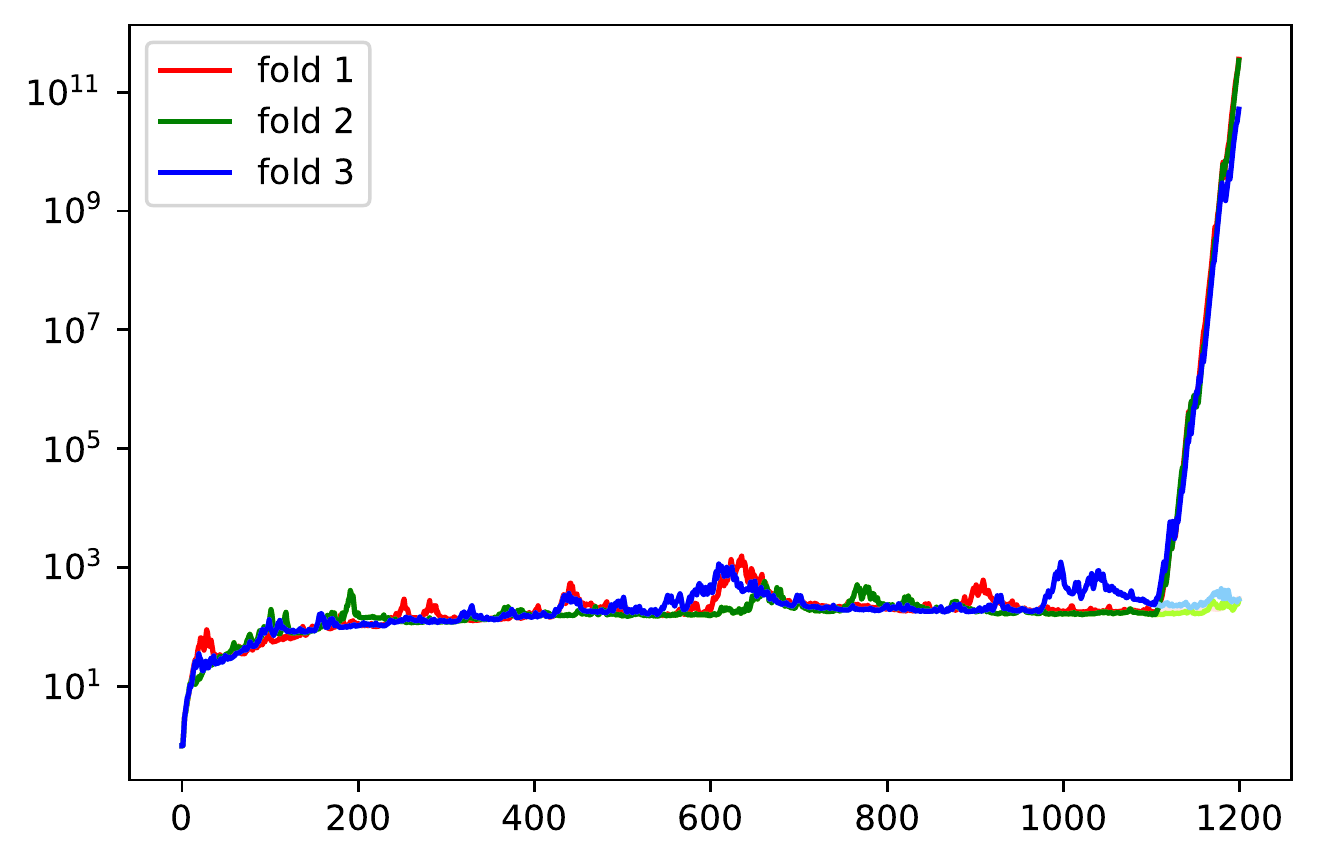}
  \end{center}
  \caption{The Shiryaev--Roberts statistic over the first 1200 observations of the combined calibration and test sets
    for Multilayer Perceptron and the conformity measure $y-\hat y$}
  \label{fig:wine-MLPdiff-SR-1200}
\end{figure}

\begin{figure}
  \begin{center}
    \includegraphics[width=\picturewidth]{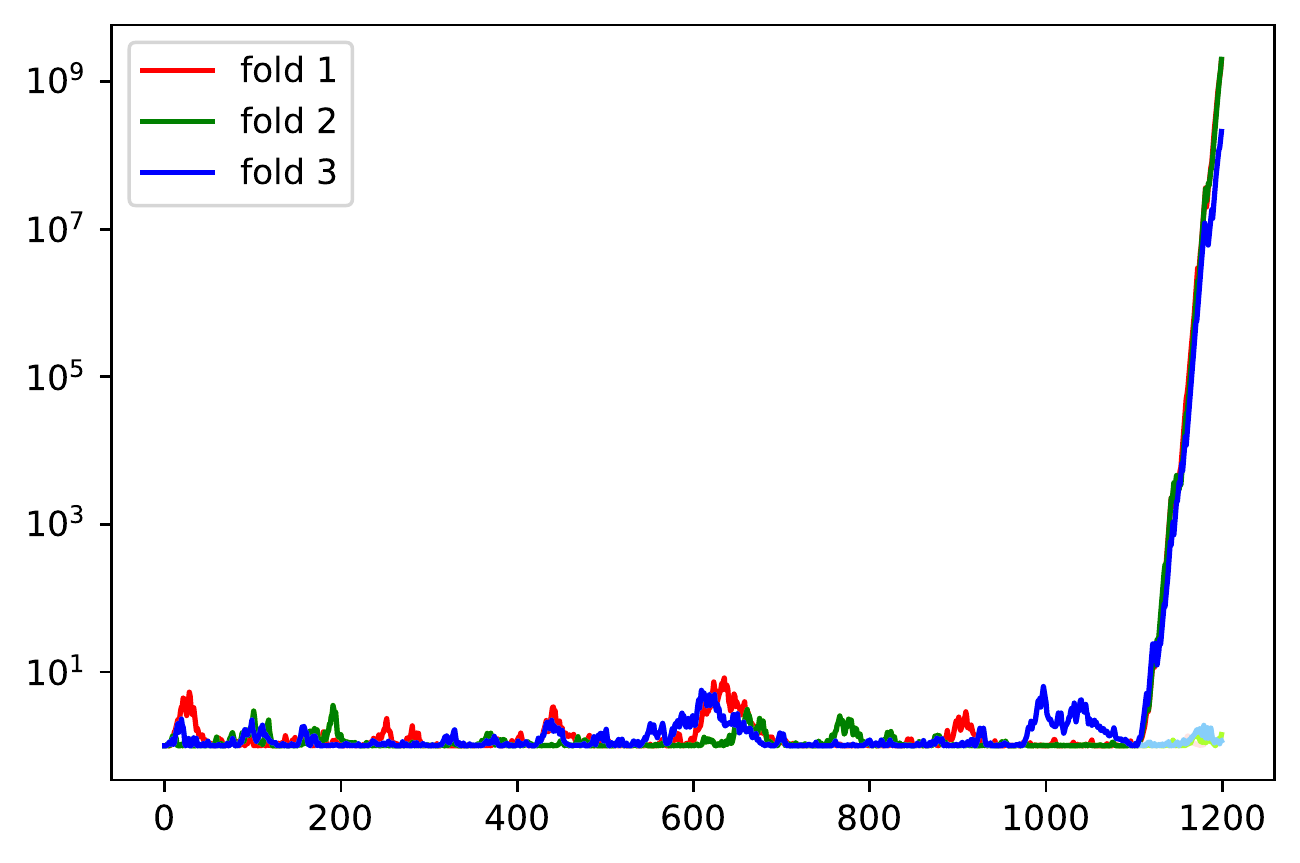}
  \end{center}
  \caption{The analogue of Figure~\ref{fig:wine-MLPdiff-SR-1200}
    for CUSUM (in the form $\gamma_n\vee1$) in place of Shiryaev--Roberts}
  \label{fig:wine-MLPdiff-CUSUM-1200}
\end{figure}

Figures~\ref{fig:wine-MLPdiff-SR-1200} and~\ref{fig:wine-MLPdiff-CUSUM-1200}
show the evolution of the Shiryaev--Roberts and CUSUM statistics
over the calibration set and the first 100 observations of the test set (in scenarios 1 and 0)
for the conformity measure $y-\hat y$, $\hat y$ being computed by Multilayer Perceptron.
Both statistics will raise an alarm in scenario 1 for $c$ up to $10^9$.

Let us now compare the performance of different prediction algorithms and conformity measures
for change-point detection more systematically,
still concentrating on the Wine Quality dataset.
The betting martingale is, as before, the Simple Jumper (with $J:=0.01$).
Our results will be summarized in Table~\ref{tab:alarms}.

\begin{table}
  \begin{center}
  \begin{tabular}{cccc}
    c.~measure & Ville & CUSUM & SR \\
    \hline
    $y-\hat y$, RF & 70 $[56,87]$ & 68 $[57,85]$ & 64 $[52,80]$ \\
    $y-\hat y$, 1NN & 92 $[69,138]$ & 93 $[70,137]$ & 86 $[65,126]$ \\
    $y-\hat y$, MLP & 55 $[44,72]$ & 55 $[45,72]$ & 51 $[42,67]$ \\
    $y-\hat y$, SVR & 156 $[100,294]$ & 156 $[102,297]$ & 142 $[94,270]$ \\
    $\left|y-\hat y\right|$, RF & 222 $[131,538]$ & 221 $[130,501]$ & 203 $[119,450]$ \\
    $\left|y-\hat y\right|$, 1NN & 192 $[114,416]$ & 188 $[115,412]$ & 172 $[106,342]$ \\
    $\left|y-\hat y\right|$, MLP & 88 $[62,143]$ & 88 $[62,142]$ & 81 $[58,130]$ \\
    $\left|y-\hat y\right|$, SVR & 720 $[300,\infty]$ & 721 $[303,\infty]$ & 562 $[272,\infty]$ \\
    $F(y)$, RF & 199 $[133,375]$ & 196 $[135,373]$ & 177 $[123,333]$ \\
    ND & 29 $[26,32]$ & 29 $[27,31]$ & 27 $[25,29]$ \\
    FND & 33 $[29,37]$ & 33 $[30,36]$ & 31 $[28,33]$
  \end{tabular}
  \end{center}
  \caption{The median delay and the interquartile intervals for the delay
    for different conformity measures and different procedures for change-point detection,
    as described in text}
  \label{tab:alarms}
\end{table}

We randomly choose a subset of 1000 white wines as training set,
a disjoint subset of 1000 white wines as calibration set,
and a subset of 1000 red wines,
all three subsets randomly ordered.
We train various prediction algorithms,
labelled with the same abbreviations as in Figure~\ref{fig:wine-accuracies},
on the training set,
use the conformity measure given in the column ``c.\ measure'' to obtain a conformal test martingale,
as described earlier,
and run the Ville, CUSUM, and Shiryaev--Roberts (SR) procedures
with the thresholds $c=10^2,10^4,10^6$, respectively,
on the calibration set continued by the test set.
The alarm is raised (i.e., the threshold is exceeded) on the test set, in the vast majority of cases,
and we define the \emph{delay} as the ordinal number of the observation in the test set
at which the alarm happens.
Table~\ref{tab:alarms} reports the median delay accompanied by the interquartile intervals
for the delays (i.e., the intervals whose end-points are the lower and upper quartiles)
for ten different conformity measures (already described earlier) and 1000 simulations;
ND stands for the nearest distance conformity measure
discussed at the end of Section~\ref{sec:Ville}.
One striking feature is how much the conformity measures $\left|y-\hat y\right|$ lose
as compared with $y-\hat y$.
The nearest distance conformity measure (detecting covariate shift) is quickest in raising alarms.

\section{Our proposed procedures}
\label{sec:proposal}

In this section we propose two procedures for deciding when to retrain,
which we call the fixed training schedule and the variable training schedule.
They are based on exchangeability martingales as described in the previous sections.
Both schedules use the Ville procedure at the beginning,
but then apply different strategies for deciding when to retrain.
For simplicity, instead of free parameters we will often use specific numbers.
Fix a prediction algorithm (such as Random Forest) and a conformity measure (such as $y-\hat y$).
Training the algorithm on a dataset then gives both a predictor (trained model)
and an exchangeability martingale (conformal test martingale).

\subsection*{Variable training schedule}

Train the prediction algorithm on the full training set,
and start running the resulting predictor on the stream of test observations.
Decide on the target lifespan of the predictor, say $C$.
\begin{enumerate}
\item
  As the first step of the testing component,
  split the training set into 3 approximately equal folds,
  1, 2, and~3.
\item
  For each $k\in\{1,2,3\}$:
  \begin{itemize}
  \item
    Train the prediction algorithm on the folds different from $k$;
    the resulting predictor gives rise to a conformal test martingale $S^k$,
    as described earlier.
  \item
    Start running the conformal test martingale $S^k$ on fold $k$ (randomly permuted)
    and then on the stream of test observations.
    Run the Shiryaev--Roberts statistic $\phi^k_n:=\sum_{i=0}^{n-1}S^k_n/S^k_i$
    on top of $S^k$.
  \end{itemize}
\item
  When two out of the three martingales $S^k$ raise an alarm at level 100, $S^k_n\ge100$, retrain
  (i.e., retrain when $\left|\{k\mid S^k_n\ge100\}\right|\ge2$).
\item
  When two out of the three Shiryaev--Roberts statistics $\phi^k$ raise an alarm at level $C$,
  $\psi^k_n\ge C$, retrain.
\end{enumerate}

\begin{figure}
  \begin{center}
    \includegraphics[width=\picturewidth]{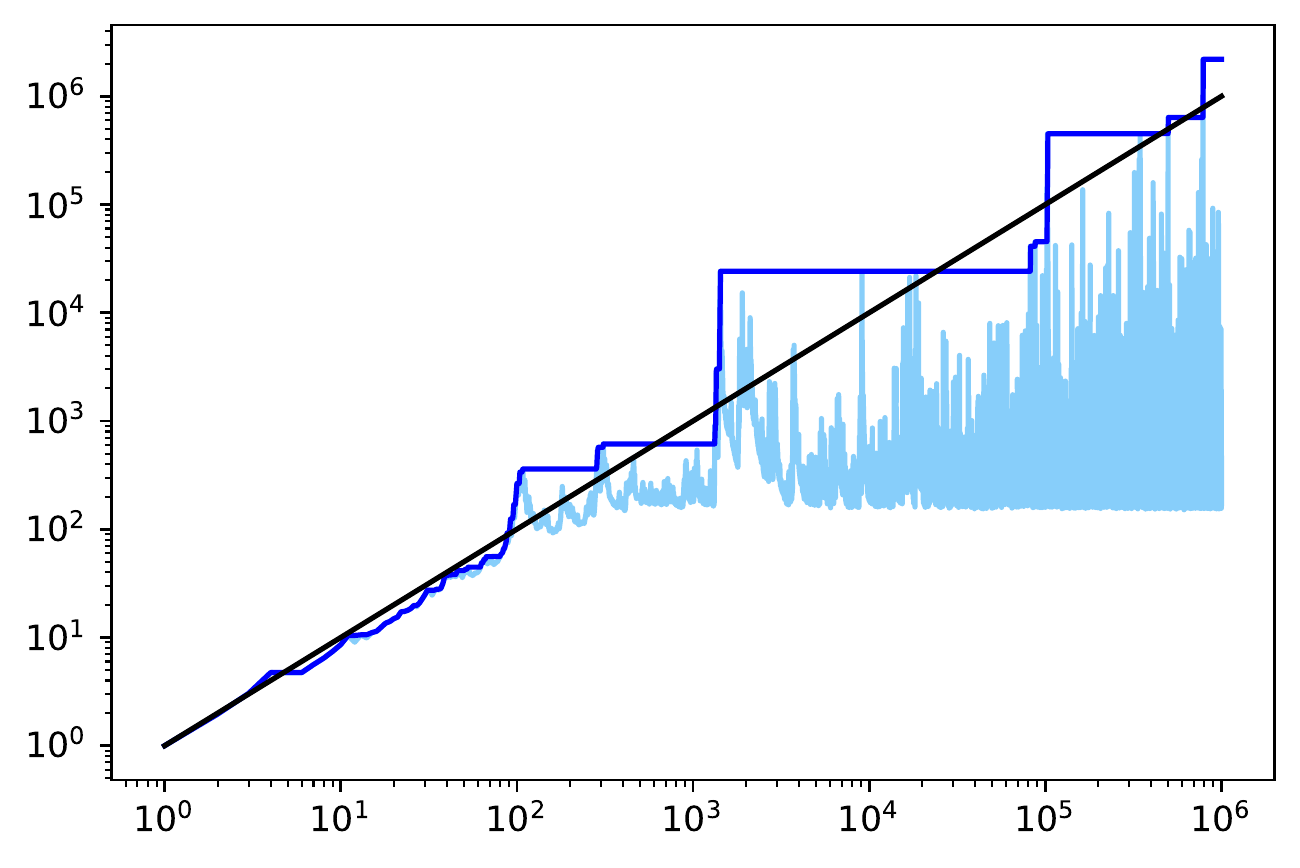}
  \end{center}
  \caption{Behaviour of the Shiryaev--Roberts statistic in the ideal setting}
  \label{fig:SR-plot}
\end{figure}

\begin{figure}
  \begin{center}
    \includegraphics[width=\picturewidth]{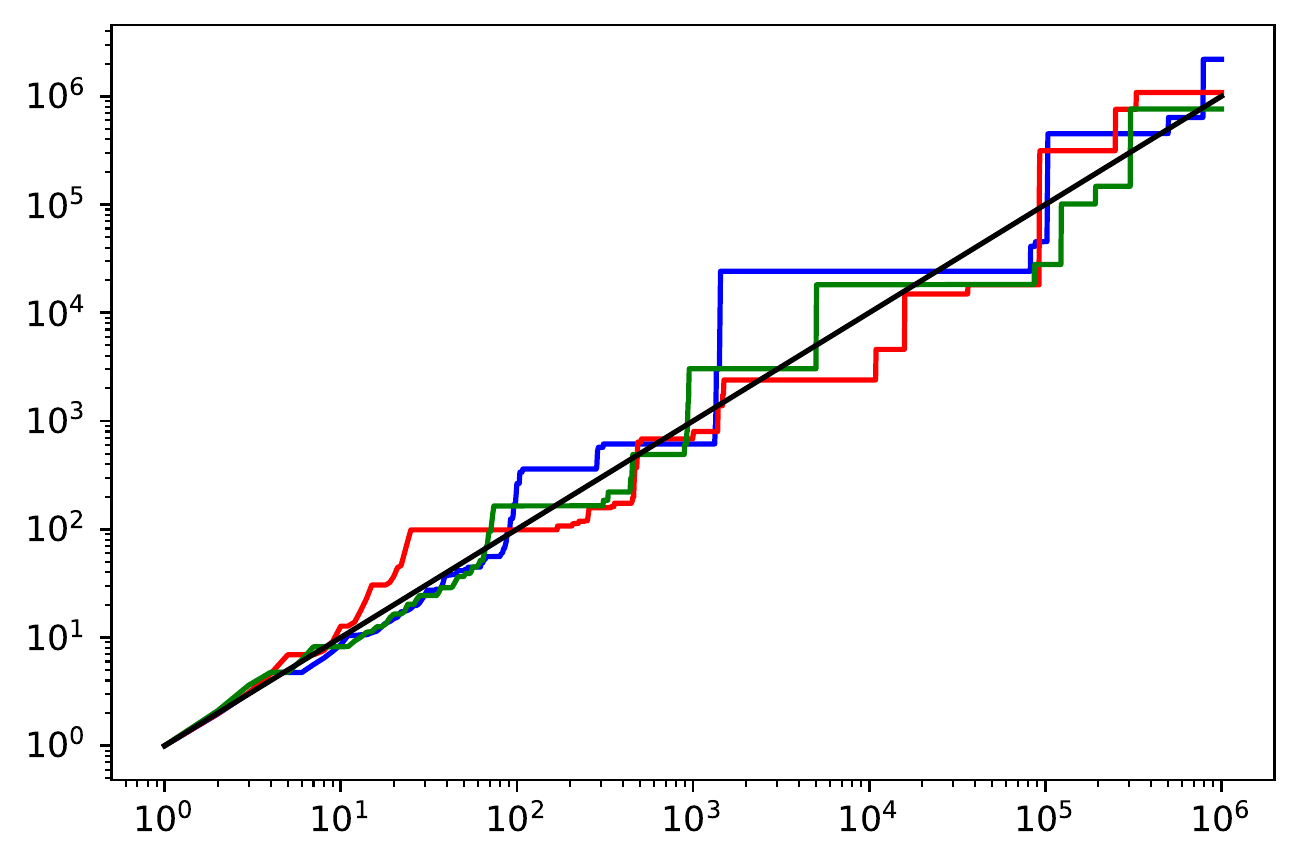}
  \end{center}
  \caption{Behaviour of the maximum process of the Shiryaev--Roberts statistic
    for three seeds of the random number generator
    in the ideal setting}
  \label{fig:SR-plots}
\end{figure}

The variable training schedule is based on the fact that $\Expect(\sigma)\ge C$
(where $\sigma$ is defined in \eqref{eq:SR})
and $\Expect(\sigma)\approx C$ when overshoots do not play a big role.
Suppose the target lifespan is $C=10^6$.
Figure~\ref{fig:SR-plot} shows the behaviour of the SR statistic $\psi_n$
(see \eqref{eq:SR}) and its \emph{maximum process}
\begin{equation}\label{eq:records}
  \psi^*_n
  :=
  \max_{i\in\{0,\dots,n\}}
  \psi_i
\end{equation}
in the \emph{ideal setting}, where the p-values are independent and uniformly distributed on $[0,1]$
(as they are under the IID assumption).
The black line is the compensator $n$ of the submartingale $\psi_n$, in the sense of $\psi_n-n$ being a martingale.
The typical behaviour of $\psi_n$ is illustrated by the light blue line in Figure~\ref{fig:SR-plot},
which is very different from its expectation, the black line.
Figure~\ref{fig:SR-plots} shows three paths of the maximum process $\psi^*_n$.

\begin{figure}
  \begin{center}
    \includegraphics[width=\picturewidth]{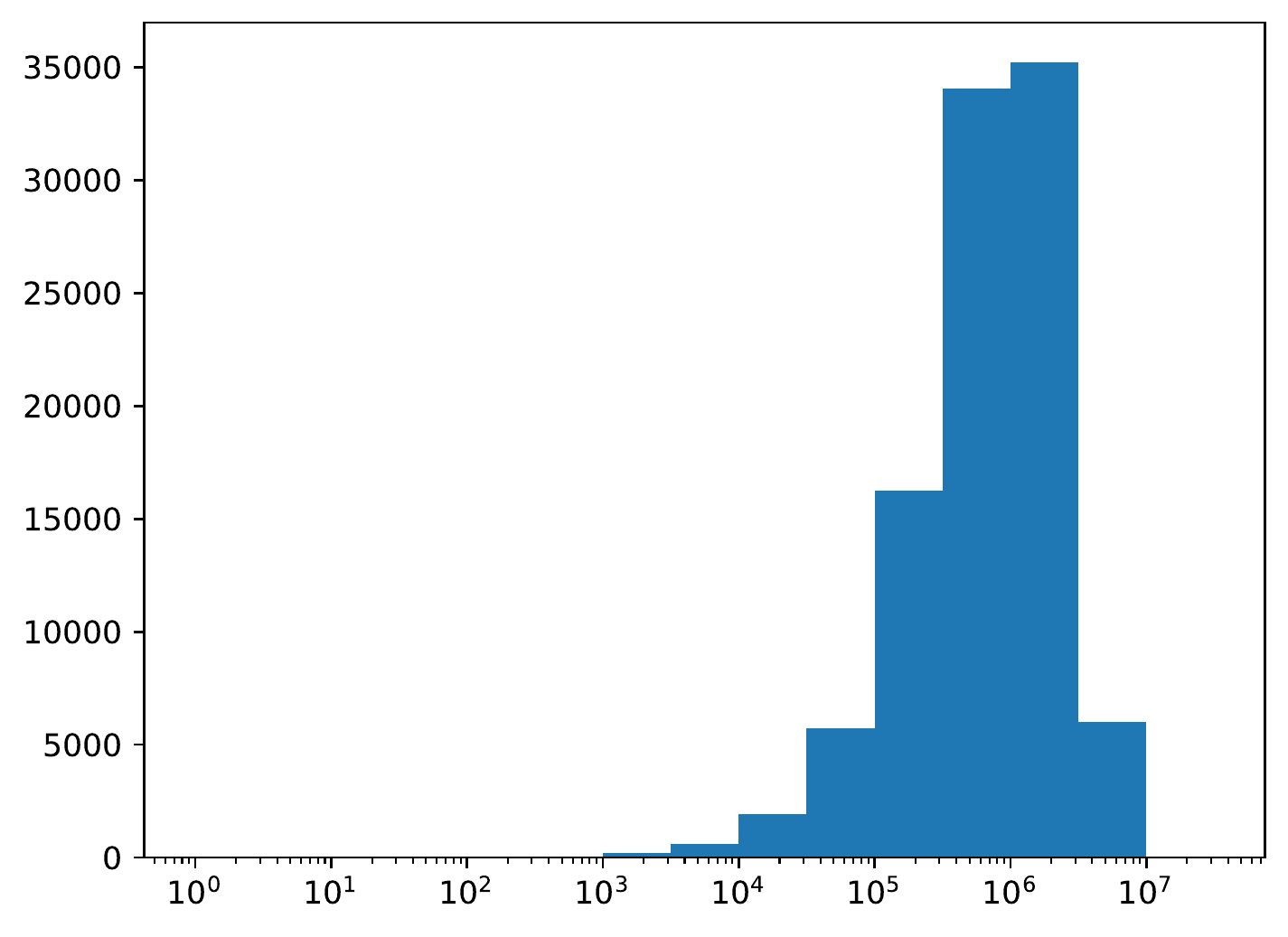}
  \end{center}
  \caption{The histogram for the time of alarm for the Shiryaev--Roberts statistic
    for the target lifespan $10^6$ and $10^5$ simulations in the ideal setting}
  \label{fig:SR-hist}
\end{figure}

Figure~\ref{fig:SR-hist} gives the histogram for the times of the alarm,
$\min\{n\mid\psi_n\ge10^6\}$, over $10^5$ simulations in the ideal setting.
In numbers, rounded to the nearest $10^4$:
the mean time to the alarm is $1.125\times10^6$, which exceeds $10^6$ because of overshoots,
with the large standard deviation of $1.123\times10^6$;
the median is $0.781\times10^6$, and the interquartile interval is $[0.320\times10^6,1.561\times10^6]$.
These figure and numbers illustrate the property of validity of the variable training schedule:
the expected lifespan of the trained predictor is indeed around $C=10^6$.
However, the variability of the lifespan, even in the ideal setting, may be a problem in some applications,
if retraining is a complicated process that needs to be planned in advance.

\subsection*{Fixed training schedule}

In the fixed training schedule we decide in advance when we would like to retrain our predictor,
and change our plans only when we have significant evidence that the distribution of the data has changed
(presumably, this will be a rare event).
Since we do not have any theoretical property of the form $\Expect(\tau)\approx C$
(the inequality $\Expect(\tau)\ge C$ still holds, of course, but it is very conservative),
our property of validity for the fixed training schedule will be computational.

\begin{table}
  \begin{center}
  \begin{tabular}{cccc}
    $f$ & alarms & $99.9\%$ confidence interval \\
    \hline
    $3.5\times10^5$ & 969 & $[0.87\%,1.08\%]$ \\ 
    $3.6\times10^5$ & 939 & $[0.84\%,1.04\%]$ \\ 
    $3.7\times10^5$ & 905 & $[0.81\%,1.01\%]$ \\ 
    $3.8\times10^5$ & 866 & $[0.77\%,0.97\%]$ \\ 
    $4\times10^5$ & 820 & $[0.73\%,0.92\%]$ \\ 
    $5\times10^5$ & 635 & $[0.56\%,0.72\%]$ 
  \end{tabular}
  \end{center}
  \caption{The confidence intervals, based on $10^5$ simulations,
    for the probability of false alarm in the fixed training schedule
    for lifespan $C=10^6$ and various choices of the threshold~$f$}
  \label{tab:endgame}
\end{table}

The fixed training schedule also starts from the target lifespan of the predictor, $C$.
\begin{enumerate}
\item
  Split the training set into 3 approximately equal folds, 1, 2, and~3.
\item
  For each $k\in\{1,2,3\}$:
  \begin{itemize}
  \item
    Train the prediction algorithm on the folds different from $k$
    getting a conformal test martingale~$S^k$.
  \item
    Start running the conformal test martingale $S^k$ on fold $k$ (randomly permuted)
    and then on the stream of test observations.
    Run the CUSUM statistic $\gamma^k_n:=\max_{i<n}S^k_n/S^k_i$
    on top of each $S^k$.
  \end{itemize}
\item\label{it:Ville}
  When two out of the three martingales $S^k$ raise an alarm at level 100, $S^k_n\ge 100$, retrain.
\item\label{it:CUSUM}
  When two out of the three CUSUM statistics $\gamma^k$ raise an alarm at level $f=f(C)$,
  $\gamma^k_n\ge f$, retrain.
\end{enumerate}
The value $f=f(C)$ should be chosen in such a way that the probability of one CUSUM statistic reaching level $f$
should not exceed $1\%$ in the ideal setting.
Let us consider, for concreteness, $C=10^6$.
One possibility is to set $f$ to the 99th percentile,
over a large number $K$ of simulations,
of the empirical distribution function
of the maximum attained by the CUSUM statistic over the random path of the Simple Jumper
between 0 and $10^6$ in the ideal setting.
Setting $K:=10^5$, we obtain, for seed 0 of the NumPy random number generator,
$3.4798\times10^5$ as the 99th percentile.
To ensure the validity of $f$,
we have computed the exact confidence intervals \citep{Clopper/Pearson:1934}
for several round values for $f$ at confidence level $99.9\%$
(to allow for multiple hypothesis testing, as we are looking at several candidates for $f=f(C)$).
For each of those $f$, we computed the number of the paths of $\gamma_n$
that trigger an alarm (which is a false alarm, since we are in the ideal setting)
at level $f$ over $n=1,\dots,10^6$;
these numbers are given in the column ``alarms'' in Table~\ref{tab:endgame}.
The confidence intervals are computed using the R package \texttt{binom} \citep{binom}.
For example, according to Table~\ref{tab:endgame}, we can set $f:=4\times10^5$,
since the corresponding confidence interval is a subset of $[0,1\%]$.
(It is sometimes argued that the Clopper--Pearson confidence intervals are too conservative
and less conservative approximate intervals are desirable,
but in our current context there is no need to sacrifice exact validity
since the number of simulations is under our control.)

The overall probability of the fixed training schedule raising a false alarm is at most $3\%$.
This follows from the Ville component \ref{it:Ville} raising a false alarm with probability at most $1.5\%$
and the CUSUM component \ref{it:CUSUM} raising a false alarm with probability at most $1.5\%$.

\subsection*{Practical aspects}

For both schedules, the predictions provided to the users of our prediction algorithm
should be computed from the full training set, of course.
The predictions computed from two out of the three folds should only be used
for monitoring the validity of the IID assumption.

Not all observations on which the trained predictor is run are necessarily included in the test stream.
In general, we have a training set, a test stream, and an exploitation stream.

We should also be careful about including observations
in the test stream in order not to violate exchangeability for irrelevant reasons.
For example, new test observations can be added in randomly shuffled batches of reasonable sizes.

\section{Conclusion}
\label{sec:conclusion}

In this paper we have discussed using conformal prediction for testing exchangeability
(this is the only known way of constructing non-trivial exchangeability martingales)
and then for deciding when a prediction algorithm should be retrained.
We have not discussed the process of retraining,
which is an interesting direction of research.
A natural question is:
which part of the available data should be used for retraining?
One possible approach is to use an exchangeability martingale
(trained on recent data) backwards:
starting from the recent data,
move into the past until the martingale detects loss of exchangeability.

This paper only scratched the surface of various specific kinds of dataset shift,
such as concept shift and covariate shift.
Those kinds will require adapting the methods proposed in this paper
and developing new ones.

\subsection*{Acknowledgments}

This research was partially supported by Amazon and Stena Line.
Thanks to Emily Hector for useful comments.

\appendix
\section{Middlegame}

A weakness of both variable and fixed training schedules is that,
in detecting deviations from exchangeability,
they concentrate on the opening and the endgame of the testing process,
to use a chess metaphor.
The Ville procedure quickly loses its efficiency as the tester's capital $S_n$ becomes very small,
and the CUSUM or Shiryaev--Roberts procedures are not efficient closer to the beginning of the test stream
since the thresholds they use are comparable with the target lifespan of the predictor.
There is a danger that in the middlegame neither opening nor endgame procedures work well.

\begin{figure}
  \begin{center}
    \includegraphics[width=\picturewidth]{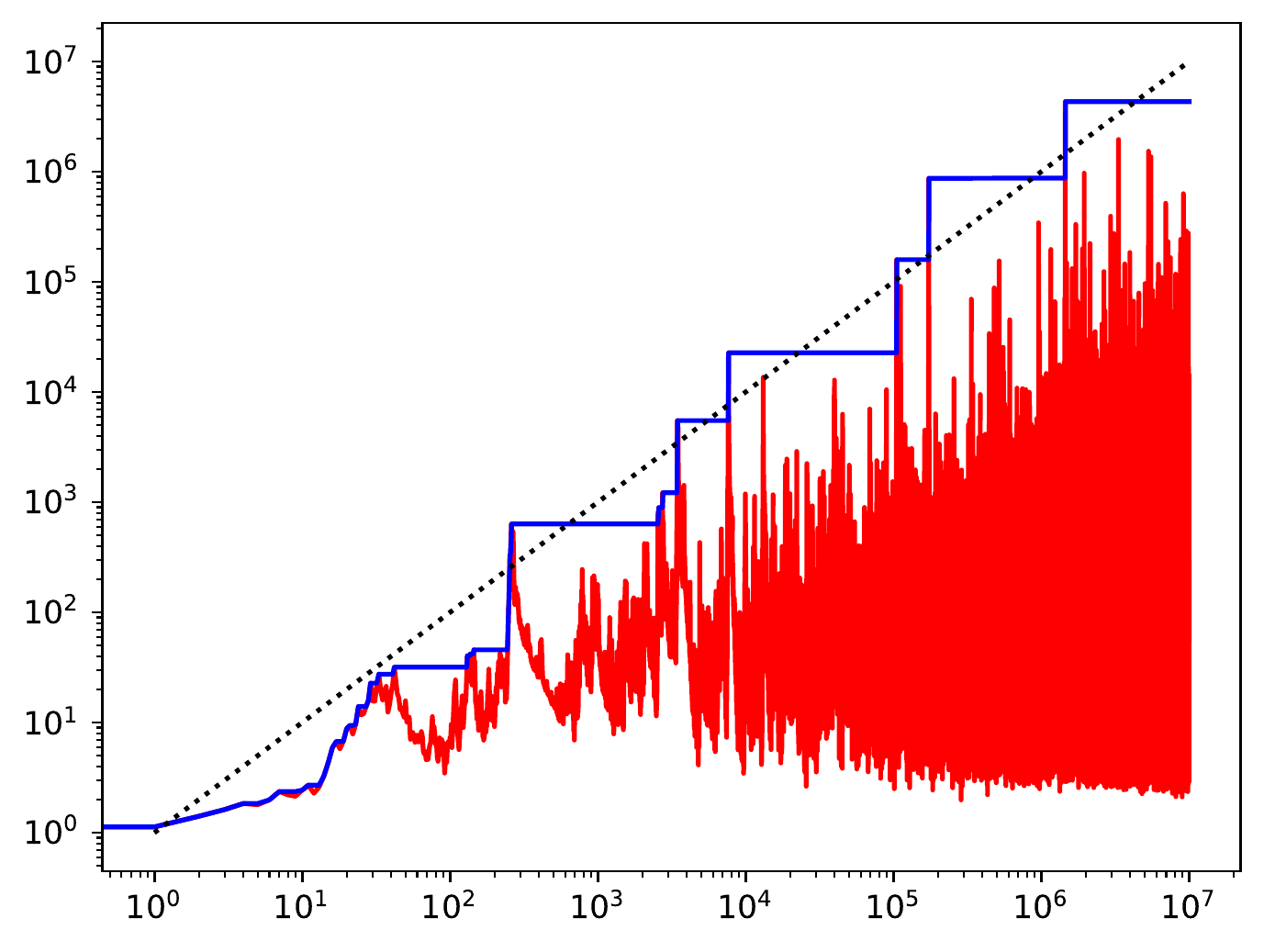}
  \end{center}
  \caption{The maximum of 100 CUSUM paths in red and its maximum process in blue
    in the ideal setting}
  \label{fig:max100}
\end{figure}

In this appendix we will discuss a procedure intermediate between the opening Ville procedure
and the endgame CUSUM procedure (we will concentrate on the fixed training schedule);
in fact, the intermediate (``middlegame'') procedure will be also a CUSUM procedure,
but the horizontal barrier implicitly used in \eqref{eq:CUSUM} and Section~\ref{sec:proposal}
will be replaced by a more complex barrier.
Figure~\ref{fig:max100} shows in red the maximum of 100 simulated CUSUM paths,
and it suggests that a reasonable barrier is a straight line with slope 1 in the loglog representation;
in the original $(x,y)$-axes the barrier has the equation $y=c x$
(a straight line passing through the origin).
The blue line in Figure~\ref{fig:max100} is defined as in~\eqref{eq:records}.

\begin{figure}
  \begin{center}
    \includegraphics[width=\picturewidth]{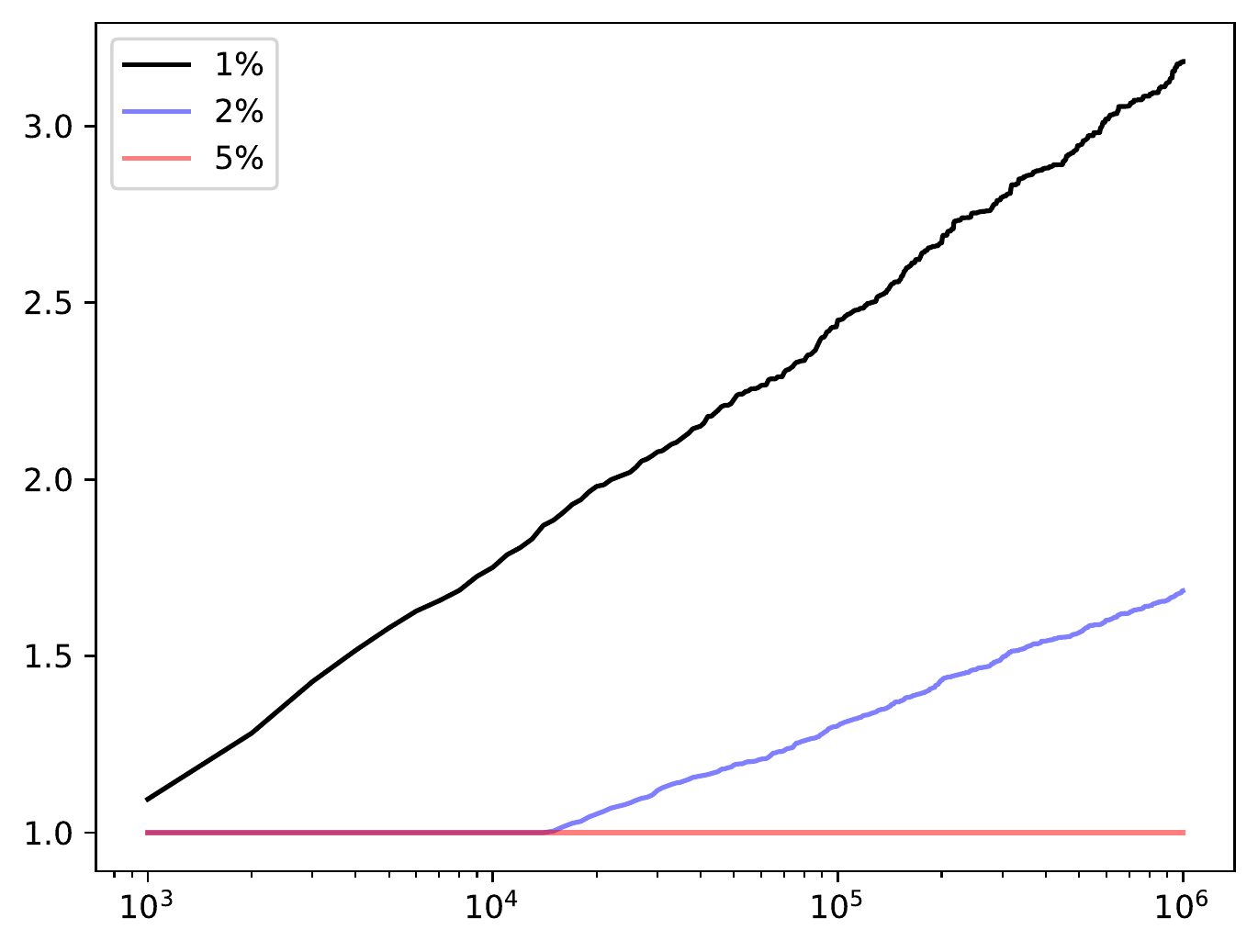}
  \end{center}
  \caption{The slopes in the middlegame in the ideal setting,
    as described in text}
  \label{fig:middlegame}
\end{figure}

\begin{table}
  \begin{center}
  \begin{tabular}{cccc}
    $c$ & alarms & $99.9\%$ confidence interval \\
    \hline
    $3.2$ & 988 & $[0.89\%,1.10\%]$ \\ 
    $3.3$ & 958 & $[0.86\%,1.06\%]$ \\ 
    $3.4$ & 930 & $[0.83\%,1.03\%]$ \\ 
    $3.5$ & 901 & $[0.81\%,1.00\%]$ \\ 
    $4$ & 793 & $[0.70\%,0.89\%]$ \\ 
    $5$ & 622 & $[0.54\%,0.71\%]$ 
  \end{tabular}
  \end{center}
  \caption{The analogue of Table~\ref{tab:endgame} for the middlegame}
  \label{tab:middlegame}
\end{table}

Figure~\ref{fig:middlegame} summarizes the empirical performance of various slopes for such barriers.
A path of the CUSUM statistic $\gamma_n$ over $n=1,\dots,N$ will trigger an alarm for a barrier $y=c x$
if $\gamma_n\ge c n$ for some $n\le N$.
Let us generate a large number $K$
($K=10^5$ in the case of Figure~\ref{fig:middlegame} and Table~\ref{tab:middlegame})
of paths of the CUSUM statistic in the ideal setting.
For each $N$, let $c_N$ be the number such that $1\%$ of the $K$ paths trigger an alarm (a false one) for the barrier $y=c_N x$.
(The definite article ``the'' in ``the number'' is almost justified for large $K$.)
The black line in Figure~\ref{fig:middlegame} plots $c_N$ vs $N$ for $N\in\{1000,2000,\dots,10^6\}$;
it looks like a straight line.
The blue line, corresponding to the $2\%$ frequency of false alarms, also looks straight, except that it cannot go under $y=1$.
This allows us to estimate the right slope $c$ for a given target lifespan of our predictor.

For example, if the target lifespan is $N=10^6$,
we can see from Figure~\ref{fig:middlegame} that $c_N\approx 3.2$ for $1\%$ 
(more precise values are $3.183$ for $1\%$ and $1.685$ for $2\%$).
To find a suitable barrier, we need a confidence interval
for the probability of a false alarm at a suitable confidence level
that is completely inside $[0,1\%]$.
According to Table~\ref{tab:middlegame}, we can take the barrier $y=4x$ (for the confidence level $99.9\%$).

Let us see how our middlegame strategy compares with the strategies that we proposed earlier
for the opening and endgame.
Suppose the target lifespan of the predictor is $10^6$, and we follow the fixed training schedule
(the comparison is easier in this case).
If we use $4\times10^5$ as the alarm threshold in the endgame (see Table~\ref{tab:endgame})
and $4 n$ as the alarm threshold at step $n$ in the middlegame (see Table~\ref{tab:middlegame}),
the former rule dominates the latter if $n>10^5$
(meaning that the former rule triggers an alarm whenever the latter does),
while the latter rule dominates the former if $n<10^5$.
We can say that, with these rules,
the middlegame ends and the endgame starts at step $10^5$.

\begin{figure}
  \begin{center}
    \includegraphics[width=\picturewidth]{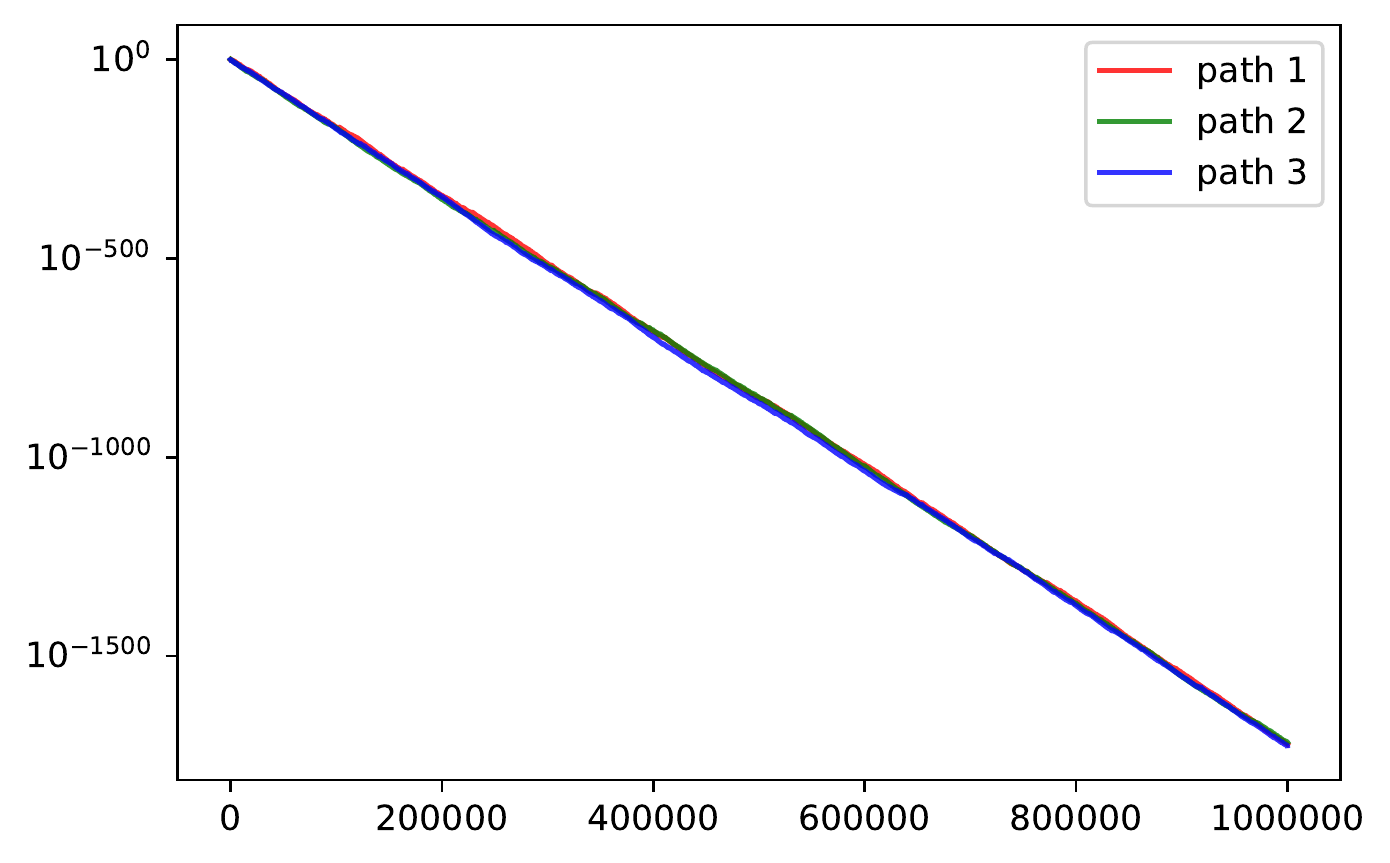}
  \end{center}
  \caption{Three typical paths of the Simple Jumper's capital in the ideal setting}
  \label{fig:Ville-paths}
\end{figure}

\begin{figure}
  \begin{center}
    \includegraphics[width=\picturewidth]{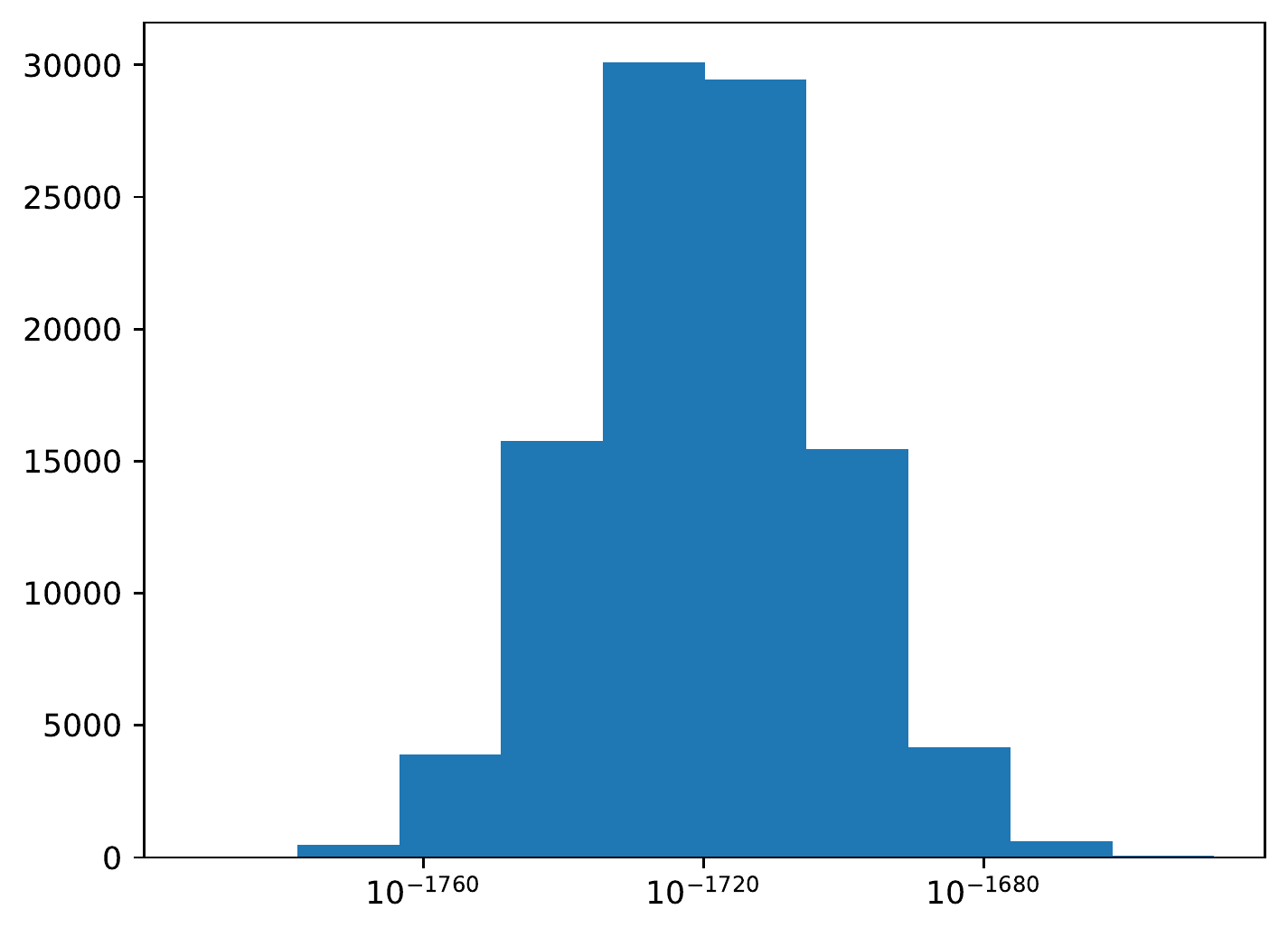}
  \end{center}
  \caption{The histogram of the final values of the Simple Jumper's capital
    based on $10^5$ simulations in the ideal setting}
  \label{fig:Ville-statistics}
\end{figure}

\begin{figure}
  \begin{center}
    \includegraphics[width=\picturewidth]{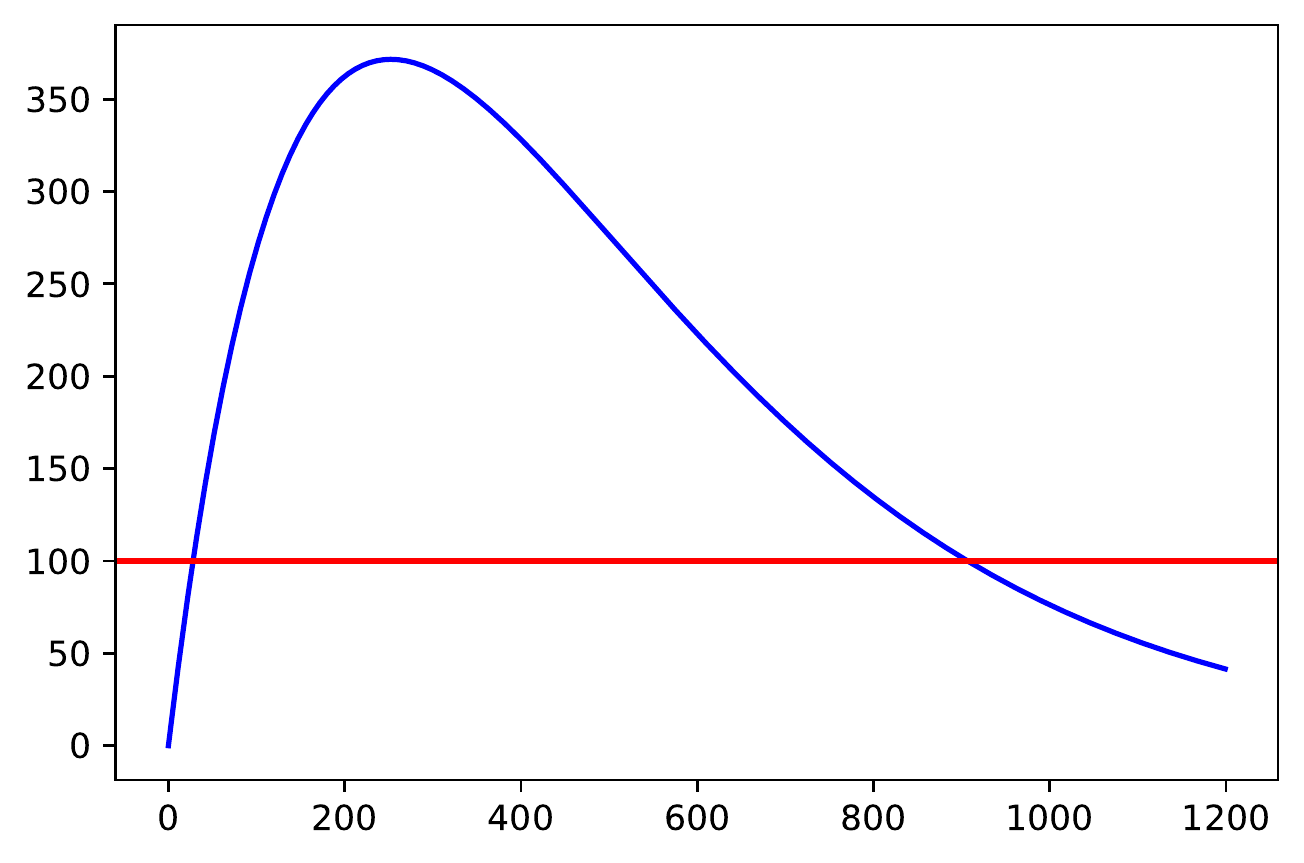}
  \end{center}
  \caption{Opening vs middlegame}
  \label{fig:equation}
\end{figure}

To compare the rules used in the opening and the middlegame,
we need to understand how the tester's capital $S_n$ evolves
during these stages of the testing process.
Figure~\ref{fig:Ville-paths} gives three typical paths
(those corresponding to the first three seeds for the NumPy random number generator)
of the Simple Jumper's capital $S_n$
in the ideal setting.
On the log-scale for the capital, they look linear and very close.
Figure~\ref{fig:Ville-statistics} is the histogram of the final values $S_{10^6}$
of the Simple Jumper over $10^5$ simulations.
In numbers, the median final capital is $10^{-1720.0}$,
and the interquartile interval is $[10^{-1731.6},10^{-1708.1}]$.
Since the Simple Jumper is a martingale,
the true expectation of its final value is 1,
but this fact is not visible in the histogram at all
(we need many more simulations for it to become visible).
If the change point is at step $n$, after which the Simple Jumper $S$ starts a quick growth,
the opening Ville procedure triggers an alarm when $S_n\ge100$,
whereas the middlegame CUSUM triggers an alarm around the time when $S_n/10^{-0.00172 n} \ge 4n$.
Solving numerically the equation $4n\times10^{-0.00172 n} = 100$,
we obtain 906.7 as its second solution
(see Figure~\ref{fig:equation},
in which the blue line is the left-hand side of the equation as function of $n$,
and the red line is the right-hand side).
Therefore, we can regard the boundary between the opening and the middlegame to be approximately $10^3$;
in the middlegame defined this way the rule of triggering an alarm when $\gamma_n \ge 4n$
can be said to dominate the Ville rule of triggering an alarm when $S_n\ge100$.

\begin{remark}
  In principle we can also use the Shiryaev--Roberts procedure in the middlegame,
  and even in the fixed-schedule endgame.
  In this paper, however, we concentrate on the more popular and intuitive CUSUM procedure.
\end{remark}

\begin{remark}
  On the other hand, we can simplify the testing process by including only the middlegame
  and discarding the opening Ville and endgame procedures altogether.
\end{remark}

\section{Avoiding numerical problems}

\begin{figure}
  \begin{center}
    \includegraphics[width=\picturewidth]{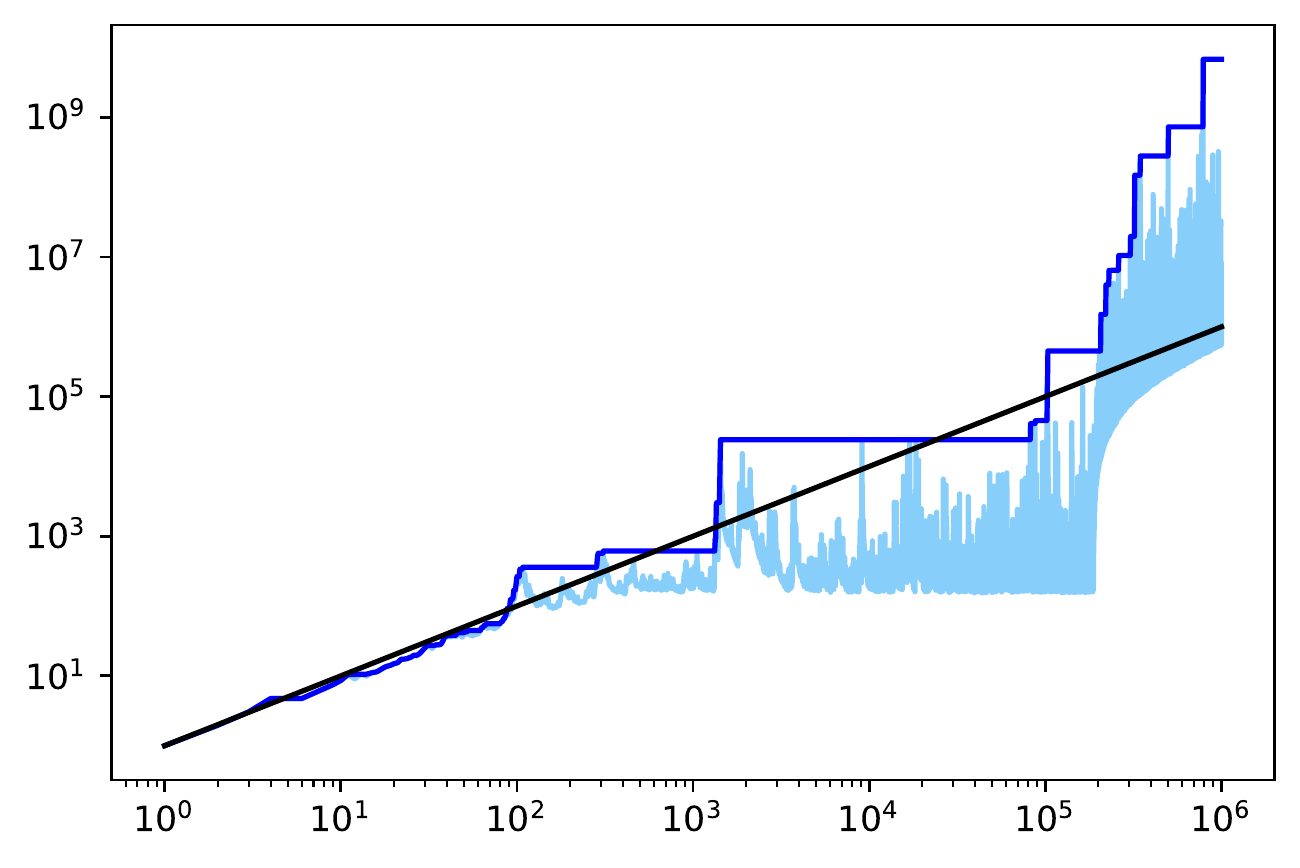}
  \end{center}
  \caption{The Shiryaev--Roberts statistic, implemented naively, over the first $10^6$ observations}
  \label{fig:SR-naive}
\end{figure}

If the Shiryaev--Roberts statistic $\psi_n$ is implemented on a modern computer directly
using the formula in \eqref{eq:SR},
we will obtain Figure~\ref{fig:SR-naive} instead of Figure~\ref{fig:SR-plot}.
The behaviour of the Shiryaev--Roberts statistic abruptly changes shortly before the 200,000th observation.
This happens because of a numerical underflow.
Up to that point the value $S_n$ of the martingale has been exponentially decreasing,
down to a small multiple of $2^{-1074}\approx5\times10^{-324}$,
the smallest positive number representable as double-precision floating-point number
(cf.\ Figure~\ref{fig:Ville-paths}).
After that point the value $S_n$ of the martingale cannot decrease further substantially
and keeps fluctuating in that small region.

To get rid of the underflow, we can use the recursion
\[
  \psi_n
  =
  \frac{S_n}{S_{n-1}}
  (\psi_{n-1} + 1)
\]
with occasional rescaling of $S_n$
(in our code we rescale $S_n$, setting it to 1, every 10,000th observation).
This results in Figure~\ref{fig:SR-plot}.
Similar precautions need to be taken for the CUSUM statistic as well,
in which case the recursion is
\[
  \gamma_n
  =
  \frac{S_n}{S_{n-1}}
  \max(\gamma_{n-1},1).
\]
\end{document}